\documentclass{article} % For LaTeX2e
\usepackage[preprint]{neurips_2021}

\usepackage{amsmath,amsfonts,bm}

\def\eqref#1{equation~\ref{#1}}
\def\1{\bm{1}}

\DeclareMathAlphabet{\mathsfit}{\encodingdefault}{\sfdefault}{m}{sl}
\SetMathAlphabet{\mathsfit}{bold}{\encodingdefault}{\sfdefault}{bx}{n}

\usepackage[utf8]{inputenc} % allow utf-8 input
\usepackage[T1]{fontenc}    % use 8-bit T1 fonts
\usepackage{hyperref}       % hyperlinks
\usepackage{url}            % simple URL typesetting
\usepackage{booktabs}       % professional-quality tables
\usepackage{amsfonts}       % blackboard math symbols
\usepackage{nicefrac}       % compact symbols for 1/2, etc.
\usepackage{microtype}      % microtypography
\usepackage{xcolor}         % colors

\usepackage{soul}
\usepackage{multirow}
\usepackage{leftidx}
\usepackage{bbm}
\usepackage{xspace}
\usepackage{color}

\usepackage{placeins}

\usepackage{enumitem}
\usepackage{graphicx}
\usepackage{caption}
\usepackage{subcaption}
\usepackage{wrapfig}
\usepackage{amssymb}
\usepackage{amsmath}
\usepackage{amsthm}
\usepackage[normalem]{ulem}
\title{Data Quality Matters For Adversarial Training: An Empirical Study}

\author{%
Chengyu Dong \\
University of California, San Diego\\
\texttt{cdong@eng.ucsd.edu}\\
\And
Liyuan Liu\\
University of Illinois at Urbana-Champaign\\
\texttt{ll2@illinois.edu}\\
\And
Jingbo Shang\\
University of California, San Diego\\
\texttt{jshang@eng.ucsd.edu}\\
}

\newcommand{\smallsection}[1]{\textbf{#1.~~~~}}

\newcommand{\specialcell}[2][c]{%
  \begin{tabular}[#1]{@{}c@{}}#2\end{tabular}}

\begin{document}

\maketitle

\begin{abstract}
    Multiple intriguing problems are hovering in adversarial training, including robust overfitting, robustness overestimation, and robustness-accuracy trade-off.
These problems pose great challenges to both reliable evaluation and practical deployment. 
Here,
we empirically show that these problems share one common cause---low-quality samples in the dataset. 
% we observe that removing low-quality samples can alleviate these problems, which implies low-quality samples to be the cause of these problems.
Specifically, we first propose a strategy to measure the data quality based on the learning behaviors of the data during adversarial training and find that low-quality data may not be useful and even detrimental to the adversarial robustness.
We then design controlled experiments to investigate the interconnections between data quality and problems in adversarial training. 
We find that when low-quality data is removed, robust overfitting and robustness overestimation can be largely alleviated; and robustness-accuracy trade-off becomes less significant.
% At the same time sounds like the same amount of data is removed for all problems.
These observations not only verify our intuition about data quality but may also open new opportunities to advance adversarial training. 
\end{abstract}

\section{Introduction}

Adversarial training~\citep{Goodfellow2015ExplainingAH, Huang2015LearningWA, Kurakin2017AdversarialML, Madry2018TowardsDL} is arguably the most effective way to establish the robustness of deep neural networks against adversarial perturbations.
Nevertheless, several intriguing problems and phenomena hover in adversarial training practice, including (1) robust overfitting~\citep{Rice2020OverfittingIA}, (2) robustness overestimation\footnote{Robustness overestimation refers to the problem that the adversarial robustness may be spuriously high against certain types of adversaries.}~\citep{Uesato2018AdversarialRA, Mosbach2018LogitPM, Croce2020ReliableEO, Chen2020RaySAR},  %\lucas{can we use spurious robustness? i feel this term sounds better...}
and (3) robustness-accuracy trade-off~\citep{Papernot2016TowardsTS, Su2018IsRT, Tsipras2019RobustnessMB, Zhang2019TheoreticallyPT}.

In this work, we focus on the data in adversarial training
% \lucas{emmm, maybe not add this sentence and ref? since it is a blog or talk, may not add lots of value} as the data-centric studies become increasingly popular in machine learning community~\citep{wu6chat} 
and show that the three aforementioned problems are all interconnected with the quality of the data employed in adversarial training. 
Specifically, we measure the data quality based on their stability of being learned through training and show that unstably learned examples are of low quality to adversarial training.
We show that low-quality data can be one common cause of the three aforementioned problems.
% in adversarial training. 
To the best of our knowledge, this is the first time the effect of the data quality being systematically studied in adversarially robust learning. 
This is also the first time that all these problems are investigated in a holistic manner. 

\begin{figure*}[t]
  \centering
  \includegraphics[width=0.98\linewidth]{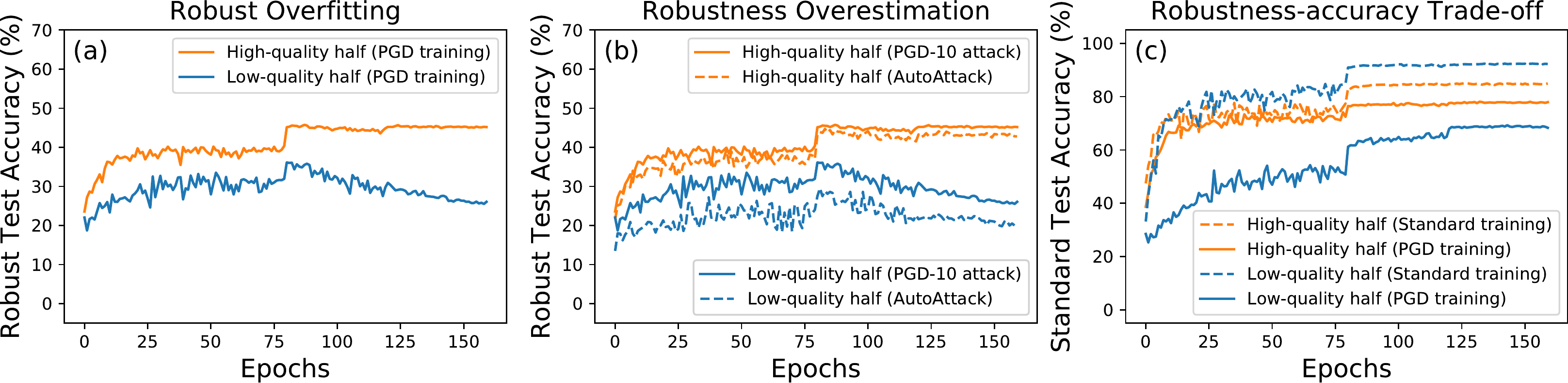}
  \caption{Despite being equal-size and class-balanced, the high-quality half and low-quality half of the CIFAR-10 training set have drastically different behaviors in adversarial training. 
  % On the high-quality half, there is only minor robustness-accuracy trade-off, almost no robust overfitting, and minimal robustness overestimation, which is in sharp contrast to the low-quality half. Here .
  %\jingbo{I feel this caption is a bit redundant to the text part. Try to use one sentence to summarize? The first sentence is a good starting point.}
  }
\label{fig:intro}
\end{figure*}

As a demonstration, we partition the training set of CIFAR-10~\citep{Krizhevsky2009LearningML} into two exclusive subsets with equal size and balanced classes, while ensuring all the examples in one subset have higher quality than all the examples in the other one in a class-wise manner.
We then conduct both standard training and adversarial training with Projected Gradient Descent (PGD)~\citep{Madry2018TowardsDL}. The model is fixed as pre-activation ResNet-18~\citep{He2016IdentityMI}.
As shown in Figure~\ref{fig:intro}, adversarial training on the high-quality half yields \emph{high} robustness, \emph{almost no} robust overfitting, \emph{genuine} robustness against the popular PGD attack, and \emph{minor} robustness-accuracy trade-off that is only noticeable in the late stage of training, which is in sharp contrast to that on the low-quality half. 

We further conduct controlled experiments on three real-world datasets, CIFAR-10, CIFAR-100~\citep{Krizhevsky2009LearningML} and Tiny-ImageNet~\citep{Le2015TinyIV}, to explore the interconnection between data quality and those three problems in adversarial training.
Our experiments cover various sample sizes, training methods (e.g., PGD and TRADES~\citep{Zhang2019TheoreticallyPT}), and neural architectures (e.g., pre-activation ResNet and Wide ResNet (WRN)~\citep{Zagoruyko2016WideRN}). 
% To further explore the interconnection between data quality and existing problems in adversarial training, we conduct extensive controlled experiments with different sample sizes, training methods including PGD training and TRADES~\citep{Zhang2019TheoreticallyPT}, neural architectures including pre-activation ResNet and Wide ResNet (WRN)~\citep{Zagoruyko2016WideRN} and datasets including CIFAR-10, CIFAR-100~\citep{Krizhevsky2009LearningML} and Tiny-ImageNet~\citep{Le2015TinyIV}. 

Our major findings are summarized as follows.
\begin{itemize}[leftmargin=*,nosep]
    \item Low-quality data may not be useful or even detrimental to adversarial training.
    \item Low-quality data causes robust overfitting in adversarial training.
    \item Adversarial training with low-quality data can lead to overestimated robustness against certain adversaries such as the PGD attack.
    \item Robustness-accuracy trade-off is only prominent when the dataset contains low-quality data. 
          When using high-quality data only, the standard accuracy achieved by adversarial training and standard training is comparable.
\end{itemize}

% \smallsection{Reproducibility}
% % We will release all implementations\footnote{\url{https://github.com/shwinshaker/RobustDataProfiling}}.
% We will release all implementations on GitHub.

The remainder of this paper is organized as follows.
In Section~\ref{sect:related}, we briefly review the existing works that are most related to our data-centric investigation of adversarial training. 
In Section~\ref{sect:pl-definition},  we propose a strategy to estimate the data quality based on learning stability, and show that unstably-learned data are of low quality to adversarial training.
Sections~\ref{sect:robustoverfitting}, \ref{sect:gradientmask} and \ref{sect:tradeoff} investigate the interconnection between data quality and major problems in adversarial training including robust overfitting, robustness overestimation, and robustness-accuracy trade-off respectively. 
Conclusions and further implications are discussed in Section~\ref{sect:conclusion}.

\section{Related work}
\label{sect:related}

Increasing attention has recently been paid to analyzing and improving the quality of the (training) data. 
Compared with standard learning, only a limited number of all learnable features would be beneficial to the adversarial robustness~\citep{Ilyas2019AdversarialEA}, thus making data quality more important in adversarially robust learning.
Existing works have been focusing on the effect of sample size and data distribution in adversarial training. 
\citet{Schmidt2018AdversariallyRG} show that adversarially robust learning requires a much higher sample complexity than standard learning. 
They also show that such sample complexity requirement is highly sensitive to the data distribution.
\citet{Ding2019OnTS} further observe that adversarial robustness achieved by adversarial training is sensitive to semantically-lossless distribution shift.
\citet{Shafahi2019AreAE} show that the adversarial robustness achieved by any classifier is fundamentally limited by the properties of the data distribution. 
Our study is distinguished from previous works by investigating the datasets with the same size and within the same distribution. We also systematically study the other important problems in adversarial training beyond the performance. 

Towards advancing adversarial robustness, substantially more works have been devoted to customizing the inner maximization or outer minimization in adversarial training based on the properties and behaviors of training data.
Representative works including \citet{Wang2020ImprovingAR}, which shows that misclassified examples have greater impacts on adversarial robustness, and \citet{Zhang2020GeometryawareIA}, which observes that some training data are more robust to adversarial attack and will significantly improve the robustness against certain types of attacks if they are emphasized in adversarial training. 
We review more works along this line in Appendix~\ref{sect:broad}. 
Our study is distinguished from these works by both methodology and objective.
In terms of the methodology, we identify the low-quality examples based on their behaviors of being learned throughout the adversarial training. 
Our approach to measure data quality yields more fine-grained results than misclassification to distinguish the individual examples, and more consistent results than the resistance to perturbation among different training methods and neural architectures (see Appendix~\ref{sect:broad}). 
In terms of the objective, we are not aiming to compete with the state-of-the-art methods, but rather to reveal the important influence of data quality on the problems in adversarial training through rigorously controlled experiments and reliable robustness evaluation. 
Such comparative and systematic analyses are lacking in the current adversarially robust learning research.

Due to the space constraint, we will discuss more works specifically related to the benefit of data in adversarially robust learning, robust overfitting, robustness overestimation, and robustness-accuracy trade-off when it comes to Sections~\ref{section:problematic-method},~\ref{sect:robustoverfitting},~\ref{sect:gradientmask} and~\ref{sect:tradeoff}, respectively. 
More broadly related work regarding data quality in standard and robust learning will be discussed in Appendix~\ref{sect:more-related}.
\section{Data quality in Adversarial Training}
\label{sect:pl-definition}

\subsection{Data Quality Measure: Learning Stability as an Example}
\label{sect:pl-estimation}

% \jingbo{You may want to make the subsection name more informative. Something like ``Learning Stability --- An intrinsic data quality measurement''}

We propose to measure the quality of the data in adversarial training based on the frequency of events that an example is predicted correctly under adversarial perturbation throughout the training, which we will refer as the \emph{learning stability} in the rest of the paper. 
Specifically, given an example $x$, the model $f$ and the optimizer $\omega$, the learning stability $s(x; f, \omega)$ can be formulated as
\begin{equation}
s(x; f, \omega) = \frac{1}{T} \left|\left\{t | f(t, x + \delta) = y(x),~t\in\{1,\cdots,T\}\right\}\right|,
\end{equation}
where $\delta$ is a $\ell_\infty$ norm-bounded adversarial perturbation, $f(t, \cdot)$ denotes the classifier at epoch $t$, $y(x)$ denotes the true label of example $x$, and $T$ is the total number of epochs the model will be trained.  
A similar measure in standard training has been utilized by \citet{Moon2020ConfidenceAwareLF} to regularize the confidence estimates of deep neural networks, and is shown to be roughly proportional to the probability that an example will be correctly predicted by a model.

\begin{figure}[h!]
    \centering
    \begin{subfigure}{0.24\textwidth}
      \centering
      \includegraphics[width=0.95\linewidth]{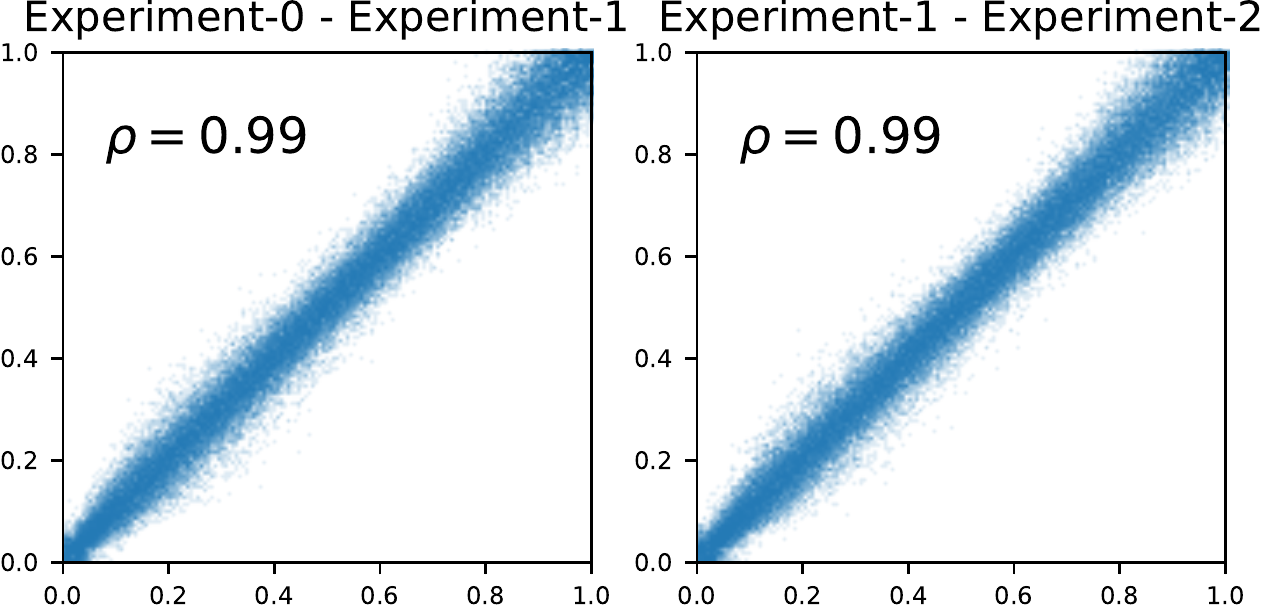}
      \caption{Random initializations}
      \label{fig:agreement-experiments-corr}
    \end{subfigure}%
    \begin{subfigure}{0.24\textwidth}
      \centering
      \includegraphics[width=0.95\linewidth]{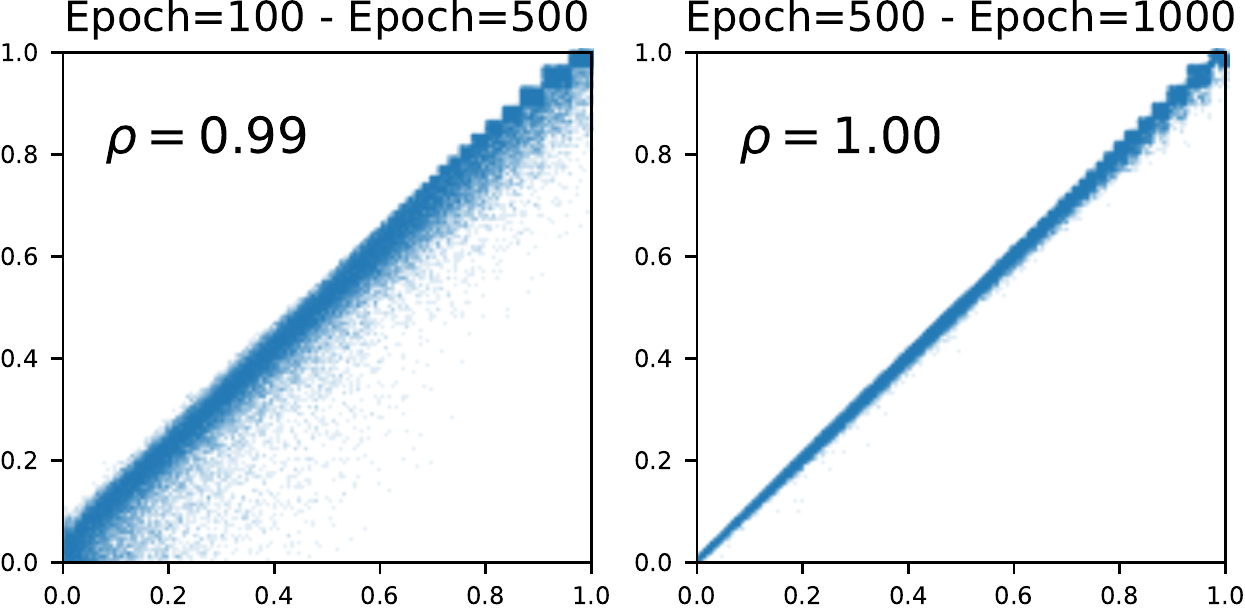}
      \caption{Training epochs}
      \label{fig:agreement-epochs-corr}
    \end{subfigure}%
    \begin{subfigure}{0.24\textwidth}
      \centering
      \includegraphics[width=0.95\linewidth]{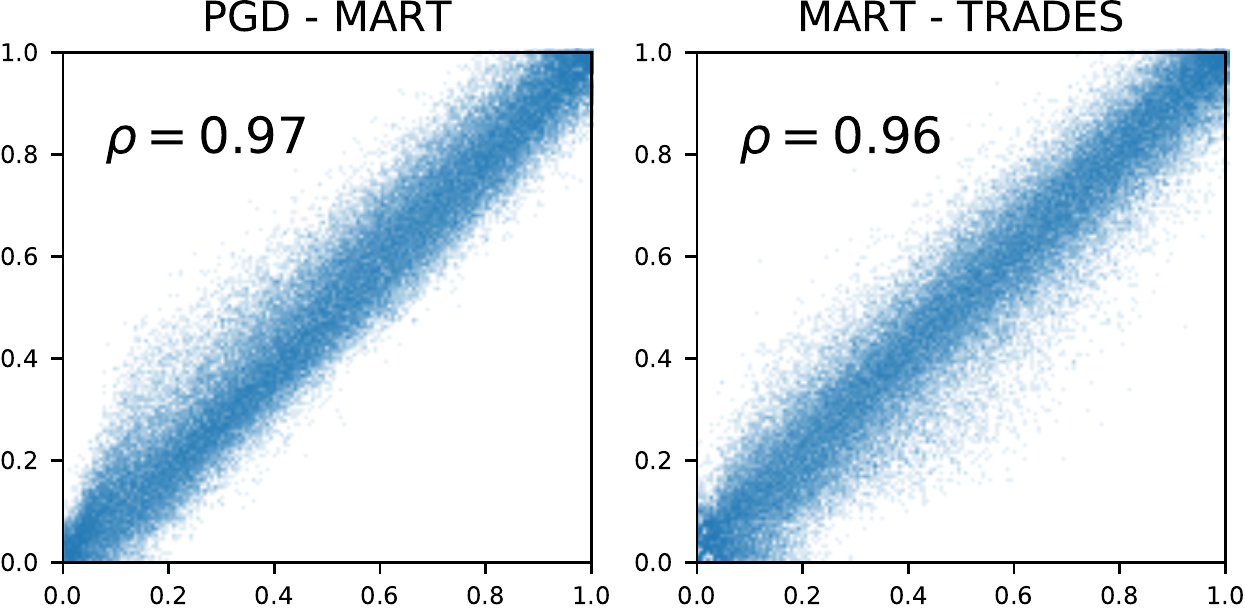}
      \caption{Different methods}
      \label{fig:agreement-objectives-corr}
    \end{subfigure}
    \begin{subfigure}{0.24\textwidth}
      \centering
      \includegraphics[width=0.95\linewidth]{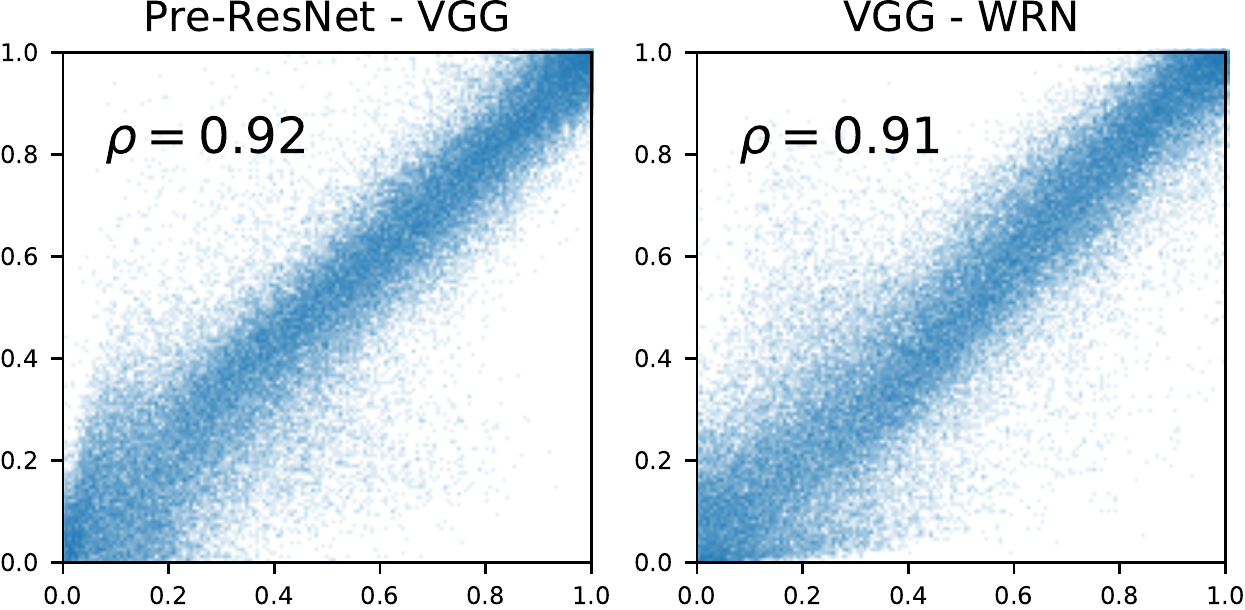}
      \caption{Different models}
      \label{fig:agreement-model-corr}
    \end{subfigure}
    \caption{Correlation between the rankings of the examples (scaled by the total number of examples) based on learning stability estimated by different training settings. Spearman's rank coefficient ($\rho$) is annotated at the upper left of each subfigure. Here we consider adversarial training methods including PGD, TRADES and MART~\citep{Wang2020ImprovingAR}, and neural architectures including pre-activation ResNet-18, VGG-19~\citep{Simonyan2015VeryDC} and WRN-16-10.}
    \label{fig:agreement-corr}
\end{figure}

We find that, although the absolute value of the learning stability can vary a lot (see Appendix~\ref{section:pl-distribution}), the ranking of examples based on the learning stability is consistent across different experimental setups, such as random initializations, training epochs, training methods, and neural architectures as shown in Figure~\ref{fig:agreement-corr}. 
This suggests that the learning stability can reflect the \emph{intrinsic} ``hardness'' of an example in a \emph{relative manner}.
% \chengyu{Shaw we use Spearman rank correlation? Maybe append on the top of the figure}
% that shared by different training settings. 
% In other words, unstably-learned examples may be intrinsically hard.

% Figure~\ref{fig:agreement-experiments-corr} shows the scatter plots of problematic ranks estimated by $3$ randomly initialized experiments, which are highly consistent with each other. 
% The problematic rank is also consistent across different adversarial training methods including PGD, TRADES\footnote{TRADES employs a different method to generate adversarial perturbation compared to PGD. To ensure a consistent definition of learning stability, we implement an additional procedure to generate adversarial perturbation with PGD-10 for TRADES.}, MART~\citep{Wang2020ImprovingAR} and different models including VGG~\citep{Simonyan2015VeryDC}, pre-activation ResNet and Wide ResNet (WRN), as shown in Figure~\ref{fig:agreement-objectives-corr} and  Figure~\ref{fig:agreement-model-corr}.

% Towards a more accurate estimation of the data quality, 
Towards a more accurate estimation of such hardness, we calculate the ranking of the learning stability based on the ensemble of multiple experiments (see Appendix~\ref{sect:pl-calculation}). 
Figure~\ref{fig:low-quality-examples} and Figure~\ref{fig:high-quality-examples} 
% show samples of the examples with the lowest quality and the examples with the highest quality, respectively, in the CIFAR-10 training set. 
show samples of the examples with the lowest learning stability and the examples with the highest learning stability, respectively, in the CIFAR-10 training set. 
One can find that unstably-learned examples appear more ambiguous and may not align well with their given labels. 
Under adversarial perturbation with relatively small size, they cannot easily be identified even by humans. This aligns with our intuition that unstably-learned examples may be intrinsically hard.
In standard learning, hard examples are often essential to the model performance, which is manifested in real-world applications such as hard example mining~\citep{Kumar2010SelfPacedLF, Shrivastava2016TrainingRO} and ablation studies on the effect of removing hard examples~\citep{Toneva2019AnES}. 
Empirical evidence can also be found in our Figure~\ref{fig:intro}(a) where standard training on the unstably-learned half yields a higher standard accuracy than that on the stably-learned half. 

In adversarial training, rather surprisingly, we observe an opposite case.
Unstably-learned examples are often not useful and may even hurt the robustness achieved by adversarial training (see Section~\ref{section:problematic-method}).
% in which sense those unstably-learned examples are of low quality to adversarial training. 
Therefore, in the rest of the paper, we refer those unstably-learned examples as \emph{low-quality examples}, and measure the data quality relatively using the learning stability.
It is worth mentioning that, in Appendix~\ref{sect:broad}, we show that it is possible to estimate the data quality based on other behaviors and properties of the data (e.g., predictive probability, minimum perturbation, and learning order) and all these estimations will yield consistent findings.

\begin{figure*}[t!]
\centering
\begin{subfigure}{0.49\textwidth}
  % \centering
  \includegraphics[width=0.97\linewidth]{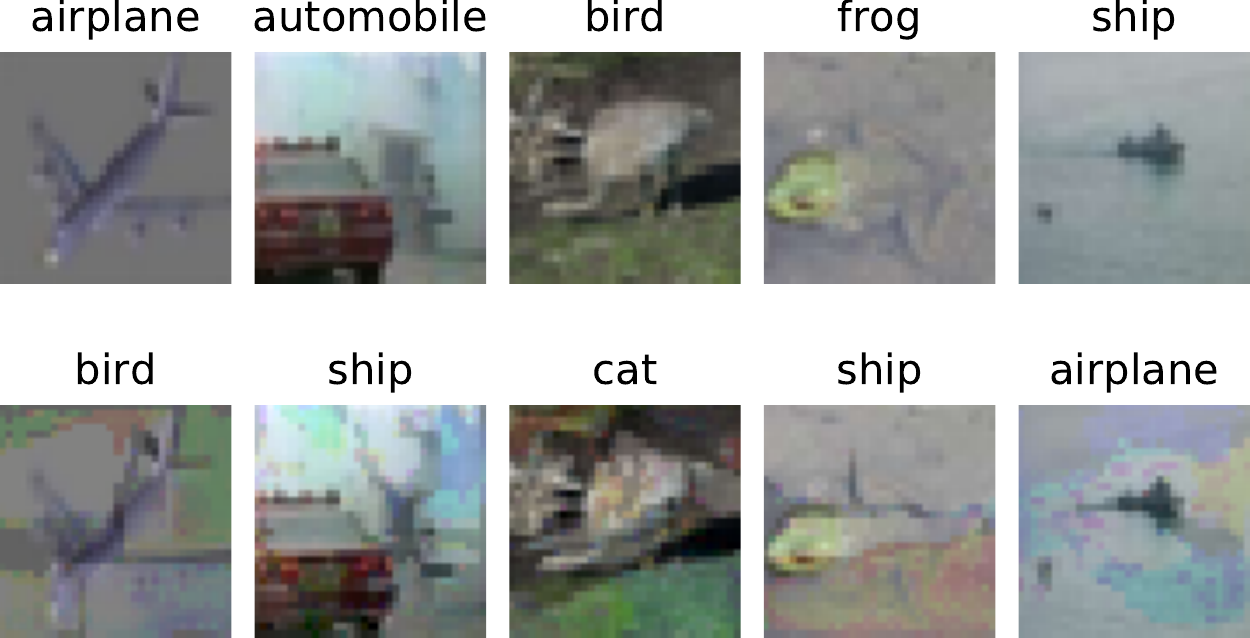}
  % \vspace{-2mm}
  \caption{Examples with the lowest learning stability}
  \label{fig:low-quality-examples}
\end{subfigure}
\begin{subfigure}{0.49\textwidth}
  % \centering
  \includegraphics[width=0.97\linewidth]{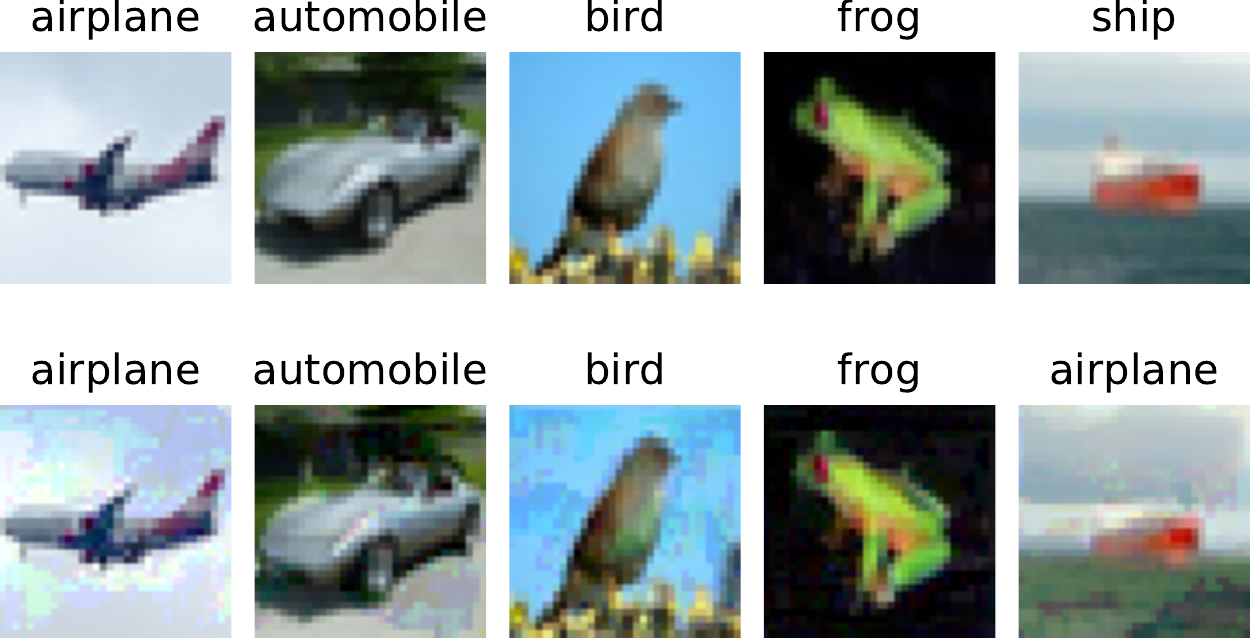}
  % \vspace{-2mm}
  \caption{Examples with the highest learning stability}
  \label{fig:high-quality-examples}
\end{subfigure}
\vspace{-2mm}
\caption{Examples selected from CIFAR-10 training set by learning stability. The top row shows the original images annotated with their true labels and the bottom row shows the corresponding adversarially-perturbed images annotated with their labels predicted by a model.}
\label{fig:examples}
\vspace{-3mm}
\end{figure*}

% \begin{figure*}[t]
% \centering
% \begin{subfigure}{1.0\textwidth}
%   \centering
%   \includegraphics[width=0.95\linewidth]{figure/examples-problematic.pdf}
%   % \vspace{-2mm}
%   \caption{Original images with true labels.}
%   \label{fig:problematic-examples-orig}
% \end{subfigure}
% \begin{subfigure}{1.0\textwidth}
%   \centering
%   \includegraphics[width=0.95\linewidth]{figure/examples-problematic-ad.pdf}
%   % \vspace{-2mm}
%   \caption{Corresponding adversarially perturbated images with predicted labels.}
%   \label{fig:problematic-examples-ad}
% \end{subfigure}
% \vspace{-2mm}
% \caption{Samples of the lowest quality in CIFAR-10 training set.}
% \label{fig:problematic-examples}
% \vspace{-3mm}
% \end{figure*}

% \begin{figure*}[t]
% \centering
% \begin{subfigure}{1.0\textwidth}
%   \centering
%   \includegraphics[width=0.95\linewidth]{figure/examples-friendly.pdf}
%   % \vspace{-2mm}
%   \caption{Original images with true labels.}
%   \label{fig:friendly-examples-orig}
% \end{subfigure}
% \begin{subfigure}{1.0\textwidth}
%   \centering
%   \includegraphics[width=0.95\linewidth]{figure/examples-friendly-ad.pdf}
%   % \vspace{-2mm}
%   \caption{Corresponding adversarially perturbated images with predicted labels.}
%   \label{fig:friendly-examples-ad}
% \end{subfigure}
% \vspace{-2mm}
% \caption{Samples of the highest quality in CIFAR-10 training set.}
% \label{fig:friendly-examples}
% \vspace{-3mm}
% \end{figure*}

\subsection{Unstably-learned data is of low-quality for adversarial training}
\label{section:problematic-method}

In this section, we conduct controlled experiments on three real-world datsets and show that low-quality examples are not useful and may even be detrimental to adversarial training.

% \chengyu{This section should serve as a verification as our claim that low-quality examples are indeed of low quality. Therefore as long as they are not that useful to the performance, or even hurt the performance would be enough. Therefore the results on other datasets can also be included.}

% \chengyu{Therefore mention more about the effect of capacity and perturbation size.}

% \chengyu{Show the problematic data is important by showing removing problematic data can help, which is significant because it clearly show the data property matters. Removing random data leads to performance drop is not insignificant.}

\smallsection{Related work}
% \jingbo{this is more like a related work section? Our point is not to argue that removing data could improve robustness. This paragraph may confuse the reviewers and lead them to challenge us.}\chengyu{here the related work is not about removing data could improve robustness}
% \jingbo{I feel this paragraph is breaking the reading flow. maybe drop it or move it to a later place. Results and discussions should come right after the experimental setup.}
% Parallel studies and findings exist in the literature showing that more data will not always yield better adversarial robustness. 
In adversarially robust learning, more data will not always yield better performance and sometimes even be detrimental.
\citet{Ding2019OnTS} observes that the robustness achieved by adversarial training plateaus after training size is sufficiently large on MNIST~\citep{LeCun1998GradientbasedLA}.
Introducing additional data unrelated to the original dataset can hurt the adversarial robustness~\citep{Uesato2019AreLR, Gowal2020UncoveringTL}. 
\citet{Yang2020RobustnessFN} shows that a proper pruning of the dataset will benefit the adversarial robustness for non-parametric classifiers. 
% \jingbo{this work seems a bit outlier here? it supports that removing data will help? It seems that the first sentence is only related to the second and third sentences? }
A recent study also shows that label noise in the dataset may be the origin of adversarial vulnerability~\citep{Sanyal2020HowBI}. 
However, we note that low-quality data is not simply label noise after manually inspecting the samples as those in Figure~\ref{fig:low-quality-examples}.
Moreover, in-depth studies show that the fractions of label noise in benchmark datasets are only marginal~\citep{Northcutt2021PervasiveLE}.
% We also find that the optimal fraction of data being removed is higher than the misclassification rate of the model at the best checkpoint against adversarial attack, which implies the problematic data might be broader than misclassified data.
% the misclassified examples in the training set is contributing to the robustness. 

\smallsection{Experimental setup of controlled experiments}
We conduct controlled experiments to dissect the effect of training data on adversarial training. Specifically, we gradually remove more examples from the training set and record the robustness achieved by adversarial training from scratch on an increasingly smaller training subset, while all other experiment settings remain the same. 
The training examples are removed either randomly, or in an ascending order of data quality, i.e., low-quality data will be removed first. 
By removing the same number of random data and low-quality data, it should be able to eliminate the effect of sample size in the comparison and reveal the effect of data quality.

We consider the robustness against $\ell_\infty$ norm-bounded adversarial attack with perturbation radius 8/255 throughout the paper. We evaluate the adversarial robustness against AutoAttack (AA)~\citep{Croce2020ReliableEO}, one of the adversarial attacks known to be relatively reliable. 
We employ popular adversarial training methods including PGD and TRADES combined with early stopping~\citep{Rice2020OverfittingIA}, which have been shown to attain the improvements of almost all other adversarial training variants~\citep{Croce2020ReliableEO, Chen2020RaySAR, pang2021bag}. 
We conduct experiments with pre-activation ResNet-18 in the main paper and Wide ResNet in the appendix. We present results on three datasets including CIFAR-10, CIFAR-100 and Tiny-ImageNet.
Other experiment details can be found in Appendix~\ref{sect:method-detail}. 
The same experimental setup will be applied to Sections 4, 5 and 6 as well.

% % \begin{figure*}[!ht]
% % \begin{subfigure}{0.48\textwidth}
% \begin{wrapfigure}{r}{0.55\linewidth} % [!ht]
% \vspace{-4mm}
%   \centering
%   \includegraphics[width=1.0\linewidth]{figure/robustness-size.pdf}
%   \caption{Best test accuracy obtained when removing different fractions of examples from the CIFAR-10 training set either randomly or in an ascent order of data quality.}
% \label{fig:robustness-size}
% \vspace{-2mm}
% \end{wrapfigure}
% % \end{subfigure}

\begin{figure*}[!ht]
% \begin{subfigure}{0.48\textwidth}
% \begin{wrapfigure}{r}{0.55\linewidth} % [!ht]
% \vspace{-4mm}
  \centering
  \includegraphics[width=0.97\linewidth]{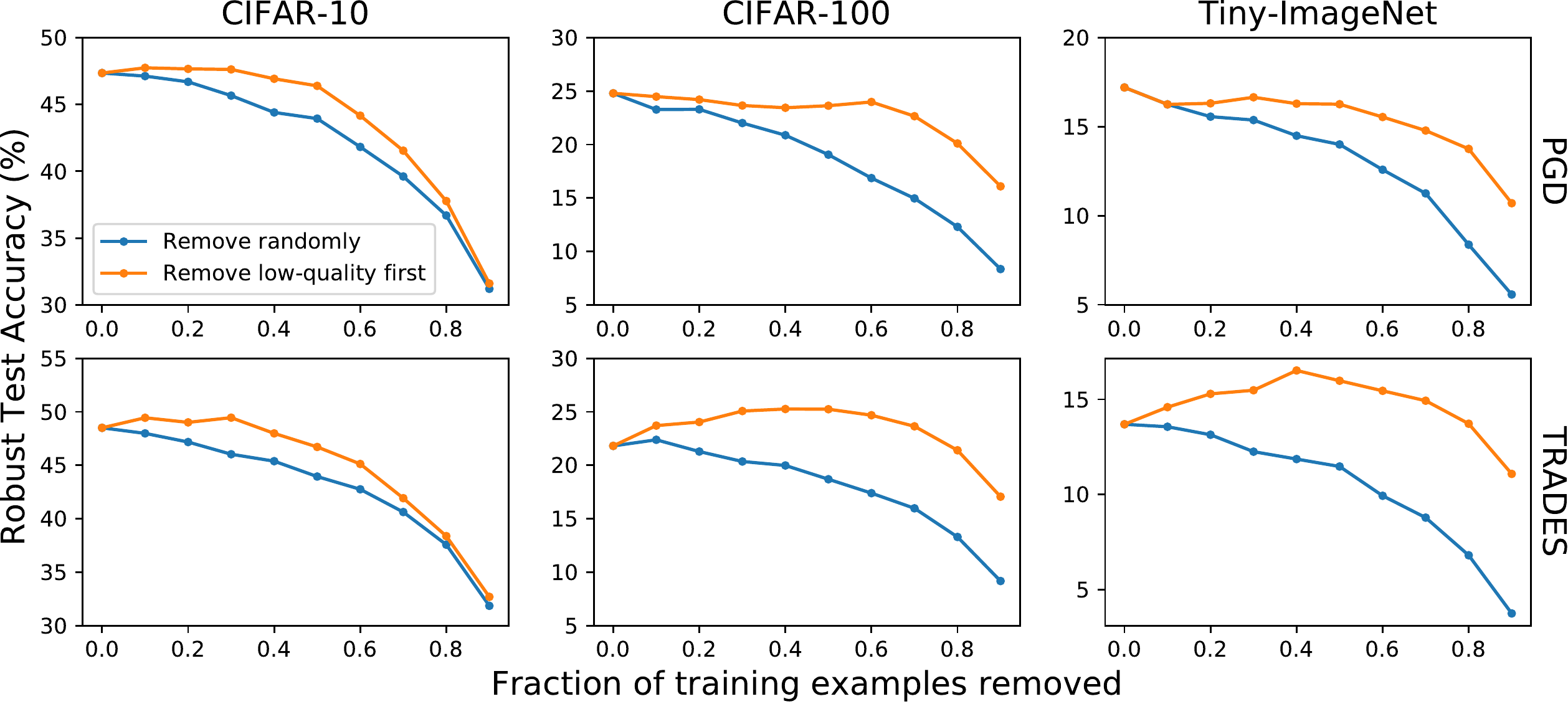}
  \caption{Best robust test accuracy obtained when removing different fractions of training examples either randomly or in an ascending order of data quality.}
\label{fig:robustness-size}
% \vspace{-2mm}
% \end{wrapfigure}
% \end{subfigure}
\end{figure*}

\smallsection{Results \& Discussions} Figure~\ref{fig:robustness-size} shows that removing random examples from the training set degrades the robustness achieved by adversarial training rapidly. 
In contrast, when removing low-quality examples first, the achieved robustness maintains or even increases.
For instance, by PGD training, up to $40\%$, $60\%$, and $50\%$ low-quality training examples can be removed without reducing the robustness significantly ($<1\%$) on CIFAR-10, CIFAR-100, and Tiny-ImageNet, respectively.
More surprisingly, removing examples with the lowest data quality can achieve robustness even higher than that achieved on the entire training set for both PGD and TRADES on CIFAR-10, and TRADES for other datasets, although the training set is actually becoming smaller.
We also validate the above observations with different perturbation sizes and neural achitectures in Appendix~\ref{append:perturbation-size-capacity}.

\section{Data Quality and Problems in Adversarial Training}

In this section, we investigate the interconnection between data quality and problems in adversarial training by conducting extensive controlled experiments.

% ==================== Robust overfitting =======================
% \subsection{Learning problematic data in adversarial training causes robust overfitting}
\subsection{Robust Overfitting}
\label{sect:robustoverfitting}
% \lucas{maybe move robust overfitting to be the first of these three sections?}
% \jingbo{I had the same quesiton before lol. I think it's a good idea to put some easier to understand things first. My order will be roubst overfitting, overestimation, and then trade-off. The trade-off section actually refers to the overestimation. }

\smallsection{The problem and related work}
% \chengyu{put the problem in the first sentence}
% \jingbo{we can drop the first sentence I think}
% In standard training, increasing model complexity induced by either larger over-parameterized models or longer training will not hurt the generalizability, which is typically referred as the ``double descent'' phenomenon~\citep{Belkin2019ReconcilingMM, Nakkiran2020DeepDD}. 
% shows that training an over-parameterized network longer is shown to not hurt the generalizability.
% However, it is observed that overfitting does occur in adversarial training~\citep{Rice2020OverfittingIA}.
\emph{Robust overfitting} is a prevalent phenomenon in adversarial training~\citep{Rice2020OverfittingIA}.
Specifically, the robust test accuracy will constantly decrease after a certain point in adversarial training, resulting in an inferior final performance. Such overfitting occurs consistently across different datasets, training settings, adversary settings, and neural architectures, and cannot be completely eliminated other than using early stopping~\citep{Rice2020OverfittingIA}. 

% In this section, we show that the robust overfitting results from the problematic examples, and the timing of its occurrence during training hinges on the learning order. % \note{We also show that adversarial training might not be appropriate for problematic examples due to their intrinsic ambiguity.}

\begin{figure}[t]
% \begin{wrapfigure}{r}{0.55\linewidth} % [!ht]
  \centering
  \vspace{-3mm}
  \includegraphics[width=1.0\linewidth]{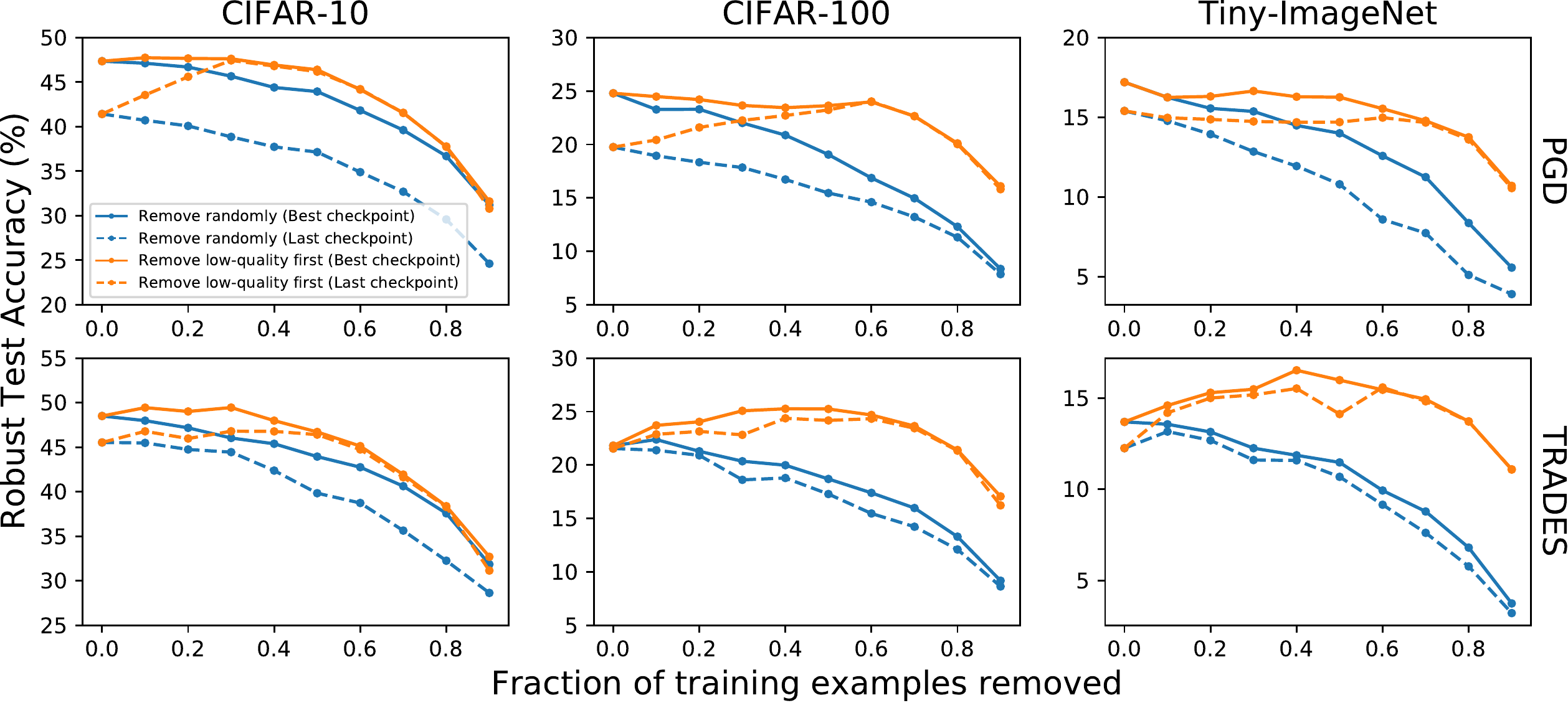}
  \caption{Robust test accuracy obtained at the best and last checkpoints by adversarial training when removing training examples either randomly or in an ascending order of data quality. 
  % The robust overfitting emerges only when the maximum problematic rank of the training set is larger than a particular threshold, implying that the problematic examples causes the robust overfitting.
  The robust overfitting can be captured by the visual gap between the dashed and solid curves of the same color.
  }
  \label{fig:overfitting-friendly-only-size}
  \vspace{-3mm}
\end{figure}

\smallsection{Low-quality data causes robust overfitting}
% We show that the robust overfitting results from the low-quality examples by our controlled experiments.
Our controlled experiments suggest that the robust overfitting results from the low-quality examples.
% , and the timing of its occurrence during training hinges on the learning order.
% \note{We also show that adversarial training might not be appropriate for problematic examples due to their intrinsic ambiguity.}
Figure~\ref{fig:intro} already shows that adversarial training on the high-quality half induces almost no robust overfitting. This suggests that robust overfitting is against our conventional understanding of overfitting in the sense that it can be mitigated by a smaller training sample size. 
Figure~\ref{fig:overfitting-friendly-only-size} further shows that when a particular fraction of low-quality examples is removed from the training set (e.g., on CIFAR-10, about $30\%$ for PGD and $50\%$ for TRADES), the robust overfitting would vanish. 
The smaller number of low-quality examples are removed from the training set, the more severe the robust overfitting will be.
In contrast, when the examples are randomly removed from the training set, the robust overfitting is often prominent consistently across different training set sizes.
This suggests that low-quality examples cause the robust overfitting, and more low-quality examples will make it more severe.

% % \begin{figure}[!ht]
% \begin{wrapfigure}{r}{0.55\linewidth} % [!ht]
%   \centering
%   \vspace{-3mm}
%   \includegraphics[width=1.0\linewidth]{figure/overfitting-friendly-only-size.pdf}
%   \caption{Best and last robust test accuracy obtained when training on various sizes of subsets either sampled randomly (``Random'') or selected based on problematic rank (``Friendly''). 
%   % The robust overfitting emerges only when the maximum problematic rank of the training set is larger than a particular threshold, implying that the problematic examples causes the robust overfitting.
%   }
%   \label{fig:overfitting-friendly-only-size}
%   \vspace{-3mm}
% % \end{figure}
% \end{wrapfigure}

% \end{wrapfigure}

% We conduct adversarial training from scratch when removing more training examples that are either selected randomly or in an ascending order of data quality.

% In Appendix~\ref{sect:more-result-overfitting}, we show that robust overfitting occurs in the late stage of the training due to the specific learning order of low-quality examples. 
% \jingbo{this is a bit off-the-topic to me. and the ``late'' is not well defined. I will suggest to remove it. learning order is kind of far from what we have so far.}

Motivated from the feature learning perspective, we conjecture that low-quality examples cause robust overfitting because robust features might be forgotten when low-quality examples are adversarially learned.
More detailed explanation can be found in Appendix~\ref{sect:explain-robustoverfitting}.

\smallsection{Mitigate robust overfitting}
Based on the above understanding, to mitigate robust overfitting, one may need to intentionally prevent the model from learning low-quality examples. 
Removing low-quality examples from the training set and the early stopping~\citep{Rice2020OverfittingIA} are the most straightforward methods to prevent such learning. 
% Beyond that, from the perspective of learning capacity, possible solutions include reducing the model complexity, adopting heavier regularization, and avoiding learning rate decay, which are partially mentioned in previous work~\citep{Rice2020OverfittingIA}. 
Recent works to mitigate robust overfitting including regularizations of the flatness of the weight loss landscape~\citep{Wu2020AdversarialWP, Stutz2021RelatingAR}, introducing low-curvature activation functions~\citep{Singla2021LowCA} and adopting stochastic weight averaging~\citep{Izmailov2018AveragingWL} and knowledge distillation~\citep{Hinton2015DistillingTK}~\citep{Chen2021RobustOM}. These methods are likely to be consistent with our data-centric understanding as various regularization techniques may suppress the learning of low-quality examples.
\subsection{Robustness overestimation}
\label{sect:gradientmask}

\smallsection{The problem and related work}
We refer \emph{robustness overestimation} as the problem that the robustness of an adversarial training method may appear spuriously high against certain types of adversaries, but be significantly diminished by stronger adversaries~\citep{Chen2020RaySAR}. Such problem might result from a significantly more difficult inner maximization induced by highly non-linear loss landscape~\citep{Qin2019AdversarialRT}, thus is more subtle than the conventional \emph{gradient masking} problem where the gradients of the model are completely not useful~\citep{Papernot2017PracticalBA, Tramr2018EnsembleAT, Athalye2018ObfuscatedGG, Uesato2018AdversarialRA, Engstrom2018EvaluatingAU}.

We focus on the PGD attack as it is the most popular method to evaluate adversarial robustness. Previous works have already shown that PGD attack with insufficient iterations~\citep{Mosbach2018LogitPM} and fixed step size~\citep{Croce2020ReliableEO} may yield misleading robustness. Existing adversarial training methods may suffer from unreliable 
% \lucas{ingeuniue seems to mean clever, maybe you mean unreliable?} \jingbo{or make it explicit as ``overly estimated''?} 
evaluation against PGD attack as mentioned in recent comprehensive studies~\citep{Croce2020ReliableEO, Chen2020RaySAR, pang2021bag}.

% \note{Tramer 2020 (On adaptive attacks...) aims to reduce the robustness of SOTA methods, can be viewed as an example of the wide existence of robust overestimation in current works}

% \smallsection{Problematic data can cause gradient masking}
\smallsection{Low-quality data can cause robustness overestimation}
For adversarial training methods, we show that low-quality data can cause overestimated robustness if evaluated against PGD attack by our controlled experiments. We report the \textbf{best robust test accuracy} throughout the training to rule out the effect of robust overfitting. 
% We show that different parts of the data can contribute to the gradient masking differently even for the same method. 
We employ AutoAttack as a stronger adversary to evaluate the adversarial robustness.
% compared to the PGD attack. %  that is widely used to evaluate the robustness. 
% We use AutoAttack to provably evaluate the robustness.
Although AutoAttack not necessarily reflects the true robustness~\citep{Tramr2020OnAA}, it is more reliable than PGD attack and is sufficient to reveal the significance of overestimation.
% gradient masking relatively.   % namely the lower bound of the robust test accuracy against any attack.

% Figure~\ref{fig:intro} already shows that adversarial training on the high-quality half can yield robustness evaluated against PGD-10 attack close to that against AutoAttack\footnote{Here (Figure~\ref{fig:intro} only) we conduct AutoAttack on a subset composed of $10^3$ randomly sampled test examples due to its heavy computational load to evaluate in every epoch.} throughout the training, which is in sharp contrast to significant gap observed from the training on the problematic partition.

% % \begin{figure}[!ht]
% \begin{wrapfigure}{r}{0.55\linewidth} % [!ht]
%   \centering
%   \vspace{-3mm}
%   \includegraphics[width=1.0\linewidth]{figure/GradientMask-size.pdf}
%   \caption{Best robust test accuracy obtained on the subsets sampled either randomly (``Random'') or selected based on problematic rank (``Friendly''). % We demonstrate the robustness overestimation by the discrepancy between the robustness evaluated against PGD-10 attack and AutoAttack. 
%   % The robustness overestimation starts to exacerbate as highly problematic data is added to the training set.
%   }
% \label{fig:overestimation-size}
% \vspace{-3mm}
% \end{wrapfigure}
% % \end{figure}

\begin{figure}[t]
% \begin{wrapfigure}{r}{0.55\linewidth} % [!ht]
  \centering
  \vspace{-3mm}
  \includegraphics[width=1.0\linewidth]{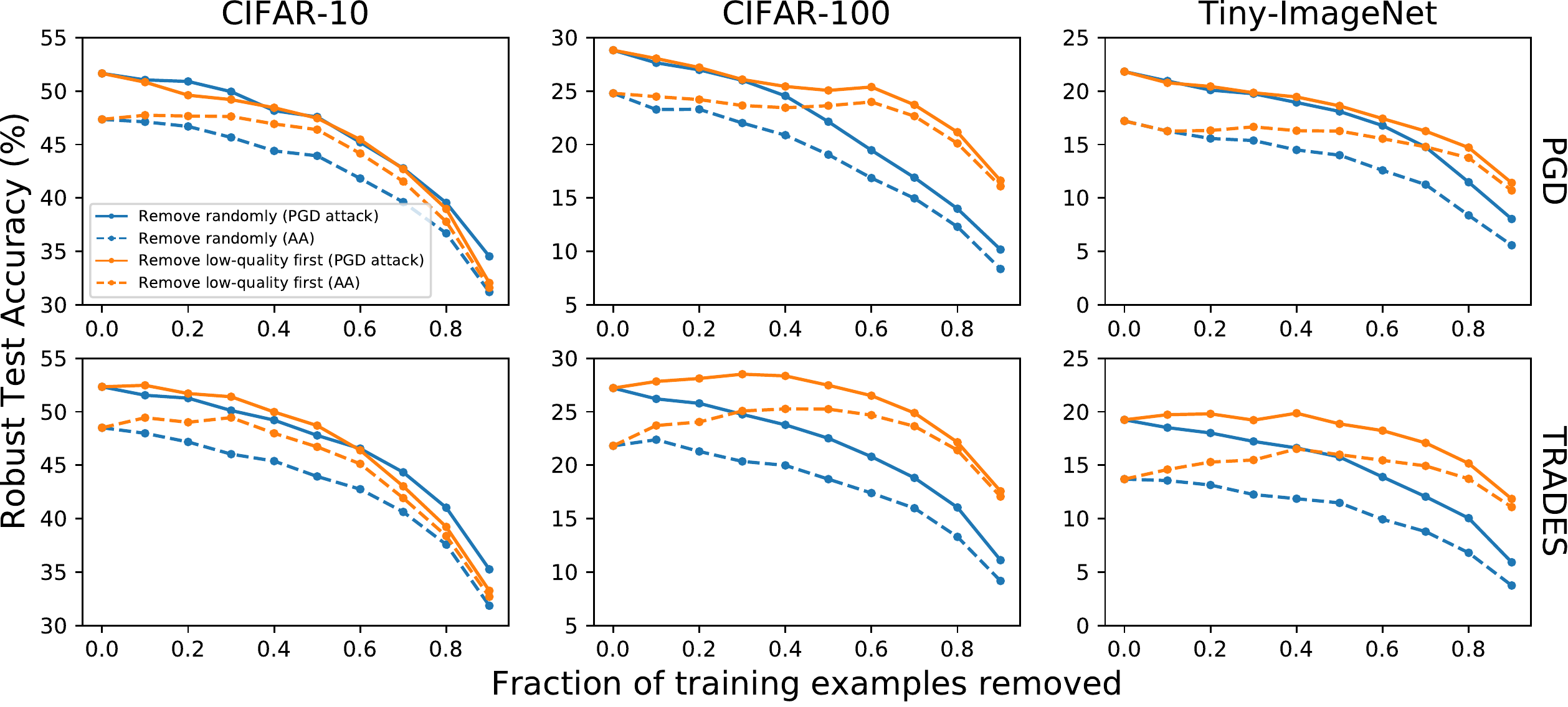}
  \caption{Best robust test accuracy obtained and evaluated against PGD attack and AA when removing training examples either randomly or in an ascending order of data quality. % We demonstrate the robustness overestimation by the discrepancy between the robustness evaluated against PGD-10 attack and AutoAttack. 
  % The robustness overestimation starts to exacerbate as highly problematic data is added to the training set.
  The overestimation gap can be captured by the visual gap between the dashed and solid curves of the same color.
  }
\label{fig:overestimation-size}
\vspace{-3mm}
% \end{wrapfigure}
\end{figure}

% We conduct the controlled experiments by removing training examples either randomly or in an ascending order of data quality. 
% \jingbo{it seems to me that we always conduct the controlled exps in the same way. Probably we can duplicate this sentence to the experimental setup.}
% sampling training subsets with different maximum problematic ranks. 

We denote the \emph{overestimation gap} as the difference between the robust test accuracy evaluated against PGD-10 attack and AutoAttack. 
% \jingbo{Again, we can mention what this gap looks like visually in the figure. It should be the gap between the dashed and solid curves of the same color.}
% Figure~\ref{fig:overestimation-size} shows that the overestimation gap expands substantially when the problematic rank of the subset approaches the upper limit (in other words, when the most problematic data is added to the training). 
Figure~\ref{fig:overestimation-size} shows that the overestimation gap shrinks significantly when low-quality data is removed from the training set. For PGD training on all datasets, the robustness obtained on the entire training set is often higher than that obtained on a smaller training set with low-quality data removed if evaluated against PGD attack, but will instead be close to or even lower than the latter if evaluated against AutoAttack. In contrast, when the data is removed randomly from the training set, the overestimation gap largely maintains and is also consistently larger than that produced by removing an equal amount of low-quality data. These evidences show that low-quality data in the training set is an important source of the robustness overestimation.

In Appendix~\ref{sect:more-result-overestimation}, we also evaluate the robustness against a variety of black-box attacks including Square Attack~\citep{Andriushchenko2020SquareAA}, RayS~\citep{Chen2020RaySAR} and transfer attack from a separately trained surrogate model, and show the robustness overestimation caused by low-quality data is consistent. 

% \note{Maybe give more intuition of geometry here as one reviewer is interested in the intuition.}
From a geometrical perspective, we suspect low-quality data causes robustness overestimation as they lie close to the decision boundary (see Appendix~\ref{sect:broad-minimum-perturbation}). The perturbation balls of low-quality data from different classes may overlap, thus twisting the local loss landscape and complicating the maximization problem in gradient-based attacks (See more discussion in Appendix~\ref{sect:explain-gradientmasking}). 

% Appendix~\ref{sect:explain-gradientmasking} shows that the low-quality data triggers robustness overestimation due to a mechanism we refer as ``competing'', a conflict of optimization targets in the inner maximization.
% \note{Should we emphasize the problematic data causing gradient masking is aligned with our common understanding, but provide a new angle from data? i.e. data are not contributing equally}

% ----------------------------------------------
\begin{table*}[htbp]
  \small
  \vspace{-1mm}
  \caption{Robust test accuracy achieved by various training methods on the entire training set, evaluated by various attack methods. ``*'' indicates the transfer attack from a surrogate model.}
  \vspace{-2mm}
  \label{table:gradient-masking-method}
  \centering
  \small
  \begin{tabular}{rcccccc}
    \toprule
    & \multicolumn{3}{c}{White-box Attack} & \multicolumn{3}{c}{Black-box Attack}\\
    \cmidrule(lr){2-4} \cmidrule(lr){5-7}
    Dataset & PGD-10 (\%) & PGD-1000 (\%) & AA (\%) & Square (\%) & RayS (\%) & PGD-1000* (\%)\\
    \midrule
    PGD & $52.07$  &  $51.04$  &  $48.12$  &  $56.19$  &  $56.86$  &  $62.24$ \\
 GAIRAT & $59.51$  &  $59.33$  &  $32.34$  &  $41.98$  &  $44.17$  &  $58.06$ \\
   Diff & $+7.44$  &  $+8.29$  &  $-15.78$  &  $-14.21$  &  $-12.69$  &  $-4.18$ \\
    \midrule
 TRADES & $52.56$  &  $51.75$  &  $48.58$  &  $55.20$  &  $55.63$  &  $63.19$ \\
   MART & $53.81$  &  $53.39$  &  $46.57$  &  $52.76$  &  $53.45$  &  $60.73$ \\
   Diff & $+1.25$  &  $+1.64$  &  $-2.01$  &  $-2.44$  &  $-2.18$  &  $-2.46$ \\
    \bottomrule
  \end{tabular}
  \vspace{-1mm}
\end{table*}

\smallsection{Implications}
Since low-quality data can cause robustness overestimation, it can be intentionally exploited to create spuriously high robustness against weak adversaries such as PGD attack. 
We find that this problem is ubiquitous among sophisticated adversarial training variants. 
We select two recently published methods GAIRAT~\citep{Zhang2020GeometryawareIA} and MART and show that they suffer from robust overestimation due to the emphases on low-quality data in their training objectives (See Appendix \ref{sect:related-overestimation}). 
As shown in Table~\ref{table:gradient-masking-method}, these methods trained with recommended settings (see Appendix~\ref{sect:method-adversary}) can achieve significant improvements compared to their corresponding baselines when evaluated against white-box PGD attack, even with a large number of attack iterations. However, when evaluating against a stronger adversary such as AutoAttack, or black-box attacks, the performance of these methods are instead inferior. Similar observations have been made for these methods in recent works~\citep{Gowal2020UncoveringTL, Hitaj2021EvaluatingTR}. These results imply that emphasizing on the low-quality data in adversarial training may not be appropriate and it is important to validate the robustness against a wide variety of evaluation metrics as more methods are focusing on customizing the training for individual examples.

\subsection{Robustness-accuracy Trade-off}
\label{sect:tradeoff}
% \lucas{maybe we can make these three sections three sub sections? would it be better?}
% \jingbo{I have tried it and please have a check. Hope it looks better.}

% \jingbo{can we have a experimental setup here to describe what are controlled, what are the variable, and what measure should the readers look at? I think this will improve the clarity significantly}
% \chengyu{Described in section 3}

% \note{setup moved to section 3}
% \smallsection{Experimental setup}
% We conduct experiments on the CIFAR-10 dataset with pre-activation ResNet-18. We employ PGD and TRADES as the adversarial training methods with $10$ attack iterations, perturbation radius $8/255$, and perturbation step size $2/255$. To reliably evaluate the robustness, we use \note{a custom version of AutoAttack} 
% (see Appendix~\ref{sect:method-detail}), unless otherwise noted. The same experiment settings are applied to Sections~\ref{sect:robustoverfitting} and \ref{sect:gradientmask} as well.

\smallsection{The problem and related work}
It is widely observed that adversarially training the model comes at the cost of standard accuracy, which is typically referred as the \emph{robustness-accuracy trade-off}~\citep{Papernot2016TowardsTS, Su2018IsRT, Tsipras2019RobustnessMB, Zhang2019TheoreticallyPT}. 
For certain learning problems, this trade-off might be inevitable either because no optimal classifier exists~\citep{Tsipras2019RobustnessMB, Zhang2019TheoreticallyPT} or the hypothesis classifier is not expressive enough~\citep{Nakkiran2019AdversarialRM}. 
Meanwhile, this trade-off is data-sensitive. 
Specifically, \citet{Ding2019OnTS} showed that the level of trade-off depends on the data distribution, and \citet{Tsipras2019RobustnessMB} observed that the adversarial training is beneficial to standard accuracy 
when the training data is insufficient to train the model. 
% \citet{Ding2019OnTS} showed that the level of trade-off depends on the data distribution. 

% In this section, we show that the situation is more nuanced. The trade-off is also sensitive to the individual data examples themselves. On friendly data, the trade-off is not significant, while on problematic data, it is prominent. 

% \chengyu{Is 'problematic data causes trade-off appropriate'?}

\smallsection{Robustness-accuracy trade-off is correlated with the data quality}
We show that the robustness-accuracy trade-off is sensitive to the data quality within the same dataset. % individual data samples themselves.
% \jingbo{better to say ``sensitive to data quality''?}
On high-quality data, this trade-off is not significant; however, on low-quality data, it is prominent.

% We show that the level of accuracy-robustness trade-off in adversarial training is positively correlated with the problematic rank of the training set. 

% % \begin{figure}[!ht]
% \begin{wrapfigure}{r}{0.55\linewidth} % [!ht]
% \vspace{-1mm}
%   \centering
%   \includegraphics[width=1.0\linewidth]{figure/tradeoff-size.pdf}
%   \caption{% Cross-generalization gap on various sizes of subsets sampled from CIFAR-10 training set. 
%   % Compared to a randomly-sampled equal-size subset, a friendly subset consisting of examples with problematic ranks below a threshold produces a consistently smaller cross-generalization gap.
%   }
%   \label{fig:tradeoff-size}
% \vspace{-1mm}
% \end{wrapfigure}
% % \end{figure}

\begin{figure}[t]
% \begin{wrapfigure}{r}{0.55\linewidth} % [!ht]
% \vspace{-1mm}
  \centering
  \includegraphics[width=0.97\linewidth]{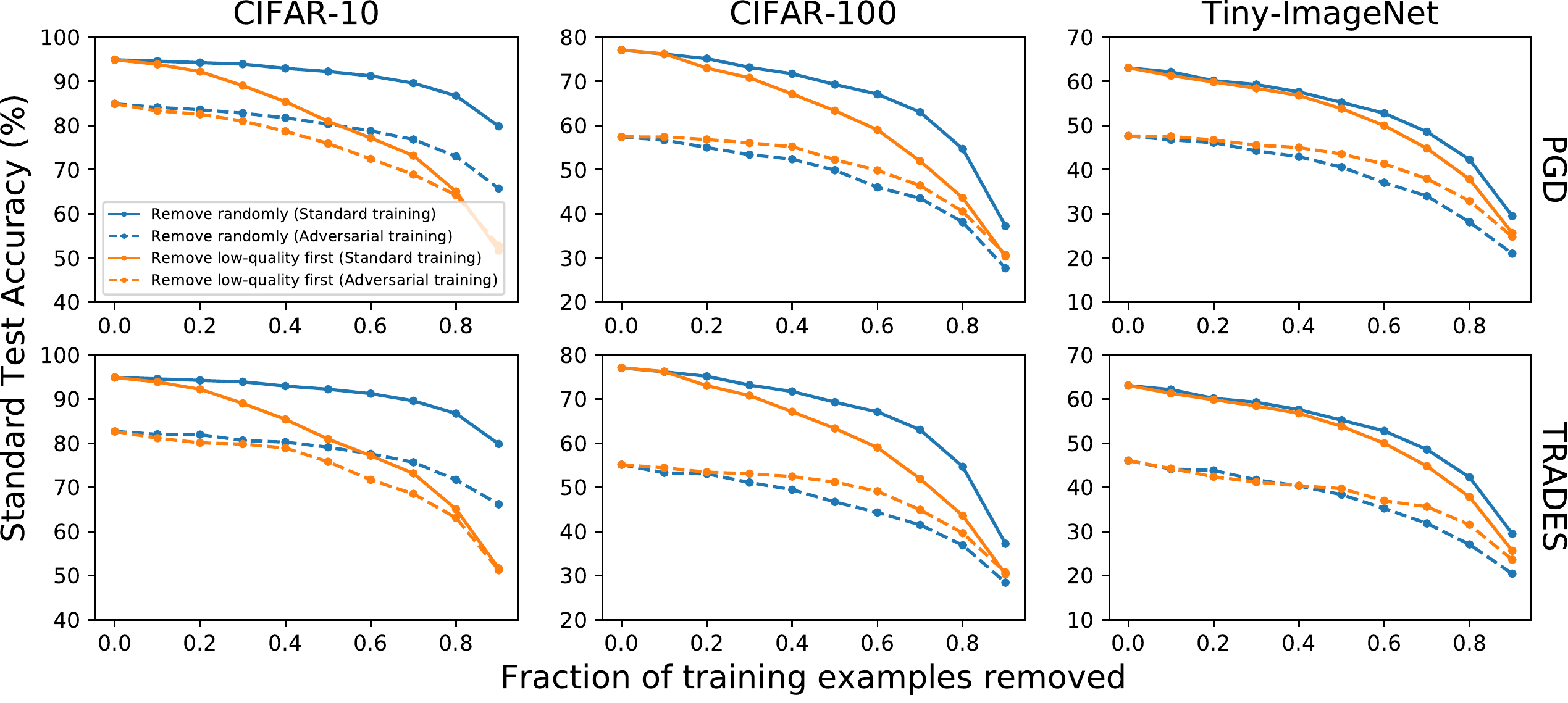}
  \caption{Standard test accuracy obtained by standard training and adversarial training when removing different fractions of training examples either randomly or in ascending order of data quality. The cross-generalization gap can be captured by the visual gap between the dashed and solid curves of the same color.
  % Compared to a randomly-sampled equal-size subset, a friendly subset consisting of examples with problematic ranks below a threshold produces a consistently smaller cross-generalization gap.
  }
  \label{fig:tradeoff-size}
% \vspace{-1mm}
% \end{wrapfigure}
\end{figure}

We present the correlation between data quality and 
% trade-off 
\emph{cross-generalization gap}
by controlled experiments.
Following previous work~\citep{Chen2020MoreDC}, we define the cross-generalization gap as 
% the difference between the standard test accuracy obtained by standard training and that obtained by adversarial training on the same dataset. 
the standard test accuracy difference between standard training and adversarial training on the same dataset, which reflects the degree of robustness-accuracy trade-off induced by adversarial training on a specific dataset. Note here we report the standard test accuracy \textbf{at the last checkpoint} for all settings in case the selected checkpoint varies due to robust overfitting.
As in Figure~\ref{fig:tradeoff-size}, the cross-generalization gap maintains the same or even enlarges as more randomly selected examples are removed from the training set. 
In contrast, the cross-generalization gap gradually shrinks as more low-quality examples are removed from the training set. 
For example, on all three datasets, adversarial training (either PGD or TRADES) and standard training yield comparable standard accuracy, when using the $10\%$ highest quality examples only (i.e., $90\%$ lowest quality ones removed).

\smallsection{Implications}
Our identified correlation between data quality and robustness-accuracy trade-off complements the existing observations of the impact of data distribution on robustness. Besides the training set size~\citep{Tsipras2019RobustnessMB} and distribution shift~\citep{Ding2019OnTS}, the quality of 
% individual characteristics of 
the training data can also influence this trade-off. 
In Figure~\ref{fig:tradeoff-size}, the cross-generalization gap induced by PGD training is negative when using the $10\%$ most high-quality examples on CIFAR-10 and CIFAR-100, which means that adversarial training is actually beneficial to the standard accuracy. 
Similar phenomena were previously only observed on MNIST or extremely small training sets ($\sim 100$ examples)~\citep{Tsipras2019RobustnessMB}.
Meanwhile, low-quality examples dominate the trade-off potentially because they are intrinsically ambiguous as shown in Figure~\ref{fig:low-quality-examples}, where robust features may be difficult to learn.
% thus hampering the model from learning their robust features. % on them. 
This is aligned with the understanding of trade-off from the feature learning perspective~\citep{Tsipras2019RobustnessMB, Ilyas2019AdversarialEA}.  % different distribution, including dimension, might not be that critical to the trade-off.
More detailed explanations motivated by this perspective will be discussed in Appendix~\ref{sect:explain-tradeoff}.

\smallsection{Mitigate robustness-accuracy trade-off}
% \note{Remove this section if no space left}
Based on the above understanding, to mitigate the robustness-accuracy trade-off one may resort to properly deal with low-quality data in adversarial training. Multiple existing works are likely to implicitly perform this by customizing either the inner maximization~\citep{Balaji2019InstanceAA, Ding2020MaxMarginA, Cheng2020CATCA, Zhang2020AttacksWD} or outer minimization~\citep{Huang2020SelfAdaptiveTB, Wang2020ImprovingAR, Zhang2020GeometryawareIA} for individual examples (see more in Appendix~\ref{sect:broad}).
Although these methods can mitigate the trade-off by increasing either robustness or accuracy while maintaining the other, the effectiveness may be unclear under state-of-the-art robust evaluation methods~\citep{Croce2020ReliableEO} as shown in Section~\ref{sect:gradientmask}. % , which is also mentioned in a recent comprehensive study~\citep{Gowal2020UncoveringTL}. 
% \chengyu{Show some results here? or just say unclear and move to future work?}
% \jingbo{Let's decide this later.}
% \jingbo{is this GAIRAT and MART against AutoAttack? If so, provide some pointers to later sections. We will discuss their effectiveness. }

% \section{Conclusions, Discussions, and Future Work}
\section{Conclusion and Discussions}
\label{sect:conclusion}
In this work we systematically study the difficulties in adversarial training from a data quality perspective. We show that the data quality is an important but previously neglected aspect for adversarial training. It is strongly correlated with the current problems in adversarial training including robust overfitting, robustness overestimation and robustness-accuracy trade-off. 

Towards understanding the difficulty of robust learning, our analyses of the data quality complements the existing analyses on the impact of data properties such as distribution and sample size. Robust learning through adversarial training can exhibit diverse levels of difficulty on individual examples in the same benchmark dataset. % the benchmark dataset itself even in the same data size regime and same data distribution. 
From human perspective, low-quality data is often ambiguous, which implies that adversarial training might be susceptible to complex and indistinct image conditions, thus challenging the feasibility of adversarial training in a realistic scenario where ambiguous data may prevail. One may note that achieving robustness through adversarial training on ImageNet~\citep{Russakovsky2015ImageNetLS} is already shown to be much harder~\citep{Rozsa2016AreAA, Kurakin2017AdversarialML}.
%%% Important to say difficutly of adversarial training, because it is this training method susceptible to the ambiguity, contrast to standard training. Other methods such as detection may not. This also shows that adversarial training exhibits more interesting `learning dynamics`.

In practice, our analyses of the data quality provides a new methodology to analyze the effectiveness of adversarial training methods. The learning behavior of a method on low-quality data reveals the potential reason why it may work or not work. For example, TRADES employs an alternative adversarial loss that penalizes the difference between the predictive probability from an adversarial example and its clean counterpart, instead of the true label, which will impose weaker supervision on low-quality examples since low-quality examples typically have low predictive probabilities (see Appendix \ref{sect:broad-probability}). As a result, TRADES can underfit low-quality data thus potentially gaining robustness compared to PGD training under the same perturbation setting, which resembles the effect of directly removing low-quality data.

Furthermore, our analyses encourage a new perspective to utilize the benchmark dataset for adversarial training. The fact that removing low-quality data maintains or even improves robustness implies that adversarial training cannot achieve the optimal robustness on the entire dataset, especially when the model capacity is relatively low. 
% Careful pruning of the training data might thus be necessary for future adversarial training practices to excel in the robustness. In addition,
A promising direction to boost the robustness is to design methods able to learn low-quality data properly such that it can also contribute to the robustness. Nevertheless, We empirically find that no existing adversarial training method, without resort to larger model capacity, can achieve reliable robustness improvement through proper learning of low-quality data. Whether or not such a method exists remains an interesting problem.

% --- The reason might lie in the fact that the adversarial training strategy itself might not be appropriate for those ambiguous examples. % === Not sure about this point because (1) Individual perturbation based on instance margin should be fine, but it actually hurts, showing the geometrical perspective might not be sufficient. (2) might because problematic data just requires higher capacity. Low capacity fitting it will only hurt (from feature perspective).
% \chengyu{Removing problematic data can benefit small model shows that the capacity of some commonly used models is still fundamentally limited in robust learning (also seen in \cite{Zhang2020GeometryawareIA})}

Lastly, as the introduction of additional labeled or unlabeled data becomes increasingly popular~\citep{Hendrycks2019UsingPC, Uesato2019AreLR, Carmon2019UnlabeledDI, Najafi2019RobustnessTA}, extra effort needs to be paid to picking high-quality data, which echos the practical suggestions made in previous works~\citep{Uesato2019AreLR, Gowal2020UncoveringTL}. % \todo{Recently a lot more works seem to emerge for additional data, cite them.}

\smallsection{Limitations}
We emphasize that removing low-quality examples only serves as a demonstration of our intuition of data quality in adversarial training, and should not be advocated as a competing method. Although removing low-quality examples can potentially advance the best robustness, it will inevitably cost standard accuracy (see Appendix~\ref{sect:performance-tradeoff}), since those hard examples are of high quality to standard training. % As noted in Section 3.2
% One may also see this from Figure 7, where removing low-quality examples for standard training reduces the standard accuracy rapidly compared to removing random examples.

% \note{Remove appendix of fairness}

\clearpage
\newpage

\section*{Reproduciblity Statement}

% We will release all implementations\footnote{\url{https://github.com/shwinshaker/RobustDataProfiling}}.
We conduct experiments on public benchmark. 
We will release implementations for all methods and scripts for all experiments on GitHub, under the Apache-2.0 license.

\section*{Ethic Statement}

In this paper, we conduct a series of controlled experiments to investigate the interconnections between data quality and problems in adversarial training. 
We experiment on three benchmark datasets that are publicly available.
We have shown that low-quality examples are helpful for standard training but may hurt the adversarial robustness.
We emphasize that removing low-quality examples only serves as a probing method to demonstrate our intuition of data quality in adversarial training, and should not be advocated as a competing method. 
Therefore, we believe our work is ethically on the right side of spectrum.

% and cannot harm any vulnerable population.

% \jingbo{We are conducting controlled experiments; We are not selling the method of removing any data. }

% Therefore, \jingbo{draw a conclusion here that our study is fair.}
% We leave the problem of how to leverage our findings to improve adversarial training as future work.

% We are aware of the fact that removing examples may bring fairness issues, which are increasingly prominent for modern machine learning systems. 
% For instance, on CIFAR-10, although removing $20\%$ low-quality data improves the robustness  by about $1\%$ for different training settings on average, it will consistently reduce the robustness achieved on the class ``Bird'' by $2\%$, which exacerbates the class disparity that may already be severe in adversarially robust learning~\citep{Nanda2021FairnessTR, Xu2021ToBR}. 
% More discussions can be found in Appendix~\ref{sect:performance-fairness}.

% \jingbo{need to draw a conclusion here.}

% \lucas{there might be ethic issue -> we conduct analyses for this -> we find no evidence that our study may hurt the ethics, instead, our finds can be leveraged to improve both the performance side and the ethics side of adversarial training. }

% \jingbo{move it to conclusion for limitation discussion.}

\bibliography{neurips_2021}
\bibliographystyle{neurips_2021}

\appendix
\section{More related work}
\label{sect:more-related}
% \chengyu{Mostly already discussed in `broad definition`}

\subsection{Robustness overestimation}
\label{sect:related-overestimation}
We discuss two recently published methods where the emphases on problematic data can be easily identified from their training objectives (See Table~\ref{table:overestimation-objective}) compared to the baseline methods.

\begin{itemize}[leftmargin=*]
    \item GAIRAT~\citep{Zhang2020GeometryawareIA} is based upon PGD training with an additional sample-wise weight in the loss function. Examples that are farther from the decision boundary will be assigned with larger weights, which amounts to an emphasis on the problematic data since the problematic score can be alternatively characterized by the distance to the decision boundary (see Appendix~\ref{sect:broad-minimum-perturbation}).

    \item MART~\citep{Wang2020ImprovingAR} can be largely viewed as a variant of TRADES with an additional term promoting the loss of those examples with low output probabilities, which also amounts to an emphasis on the problematic data based on the probability characterization of the problematic score (see Appendix~\ref{sect:broad-probability}). 
 
 \end{itemize}

\begin{table*}[!ht]
   \small
  \caption{Training objectives in the outer minimization in various adversarial training methods, where $\ell(\cdot, \cdot)$ indicates the cross-entropy loss, $f(\cdot)$ indicates the probabilistic prediction of a model and $f_y(\cdot)$ indicates the probability corresponding to class $y$, $\delta$ indicates the adversarial perturbation of $x$ generated by the inner maximization.}
  \vspace{0.5ex}
  \label{table:overestimation-objective}
  \centering
  \scalebox{1.1}{%
  \begin{tabular}{rl}
    \toprule
    Method & Training objective \\
    \midrule
    PGD & $\min_{f} \mathbb{E}_{x,y} \left[ \ell(f(x + \delta), y) \right]$ \\ 
    GAIRAT & $\min_{f} \mathbb{E}_{x,y} \left[ \boldsymbol{\omega(x, y)} \ell(f(x + \delta), y) \right]$ \\ 
    \midrule
    TRADES &  $\min_{f} \mathbb{E}_{x,y} \left[ \ell(f(x), y) + \lambda\cdot \ell(f(x), f(x + \delta)) \right]$ \\ 
    MART & $\min_{f} \mathbb{E}_{x,y} \left[ \ell(f(x + \delta), y) + \lambda\cdot \ell(f(x), f(x + \delta)) \boldsymbol{(1 - f_y(x)}) \right]$ \\
    \bottomrule
  \end{tabular}}
\end{table*}

% Indeed, compared to their corresponding baselines, we find that these two methods can achieve consistently higher training accuracy on the most problematic data throughout the training.

\subsection{Robustness-accuracy trade-off}

We note that there exists an abundant body of works that focuses on the robustness-accuracy trade-off in robust learning. In additional to the related works discussed in the main paper, here we review some works that attack this problem from other perspectives. This is by no means an exhaustive review.

% \todo{check more of citations of tsipras paper, about 495.}
It has been argued that the robustness-accuracy trade-off is inherent to the data distribution and is thus inevitable for any classifier. \citet{Tsipras2018ThereIN, Tsipras2019RobustnessMB} and \citet{Zhang2019TheoreticallyPT} theoretically show that no optimal classifier can achieve both robustness and accuracy on toy problems. \citet{Dohmatob2018LimitationsOA} formalizes this into a ``No Free Lunch'' problem, and further proves the inevitability of trade-off under mild assumptions of the data distribution. % Dohmatob2019GeneralizedNF
\citet{Nakkiran2019AdversarialRM} shows that the trade-off is inevitable because the hypothesis class is not expressive enough. \citet{Javanmard2020PreciseTI} shows that the adversarial training may improve generalization in an over-parameterized regime, but hurt it in under-parameterized regime. 

On the contrary, some works argue that the robustness-accuracy trade-off is not necessarily inevitable in a realistic setting. \citet{Raghunathan2020UnderstandingAM} shows that the trade-off stems from the over-parameterization of the hypothesis class. Robust self-training~\citep{Carmon2019UnlabeledDI, Najafi2019RobustnessTA, Uesato2019AreLR}, overcoming the sample complexity leveraging additional unlabeled data, thus can effectively mitigate the trade-off. \citet{Yang2020ACL} shows that the trade-off in practice is a result of either the model failing to impose local Lipschiztness, or not generalizing sufficiently.
\citet{Wen2020TowardsUT} and \citet{Roth2020AdversarialTI} show that adversarial training is essentially a form of operator norm regularization, thus hurting the generalizability if not properly configured.

% \cite{Schmidt2018AdversariallyRG} shows that training robust models requires more data, which can alleviate the trade-off. On the contrary, \cite{Chen2020MoreDC} shows that more data may instead expand the gap between standard and robust model.

There are more works implying that the robustness and accuracy may not be contradict by showing that adversarial examples can benefit generalization either through different perturbation generation strategies~\citep{Stutz2019DisentanglingAR} or different adversarial training strategies~\citep{Xie2020AdversarialEI}.

% \subsection{Robust overfitting}
% Robust overfitting refers to the phenomenon that the robust test accuracy will degrade after a certain point in adversarial training, which occurs across different datasets, training settings, adversary settings and model architectures. The robust overfitting is shown to be orthogonal to 'double descent' \cite{Belkin2019ReconcilingMM, Nakkiran2020DeepDD}, and cannot be completely eliminated other than early stopping the training straightforwardly. There are recent works shown to be resistant to robust overfitting \cite{Zhang2020AttacksWD, Zhang2020GeometryawareIA}, but the effectivenss of these methods are questionable under state-of-the-art robustness certification methods \cite{Croce2020MinimallyDA, Andriushchenko2020SquareAA, Croce2020ReliableEO}.

% `https://arxiv.org/pdf/2012.13628.pdf` impact of learning rate and design better scheduler.

% \subsection{Gradient masking}
% Formally gradient masking: see introduction citations
% Mostly already discussed in analysis part.

% Others addressing false robustness
% carlini 2017a: CW attack~\citep{Carlini2017TowardsET} -- this one is not included in the main paper!!
% carlini 2017b: Assess defense methods~\citep{Carlini2017AdversarialEA} 
% Carlini 2019: Seminal work for robustness guidelines~\cite{Carlini2019OnEA}

\subsection{Data profiling in standard learning}
In classical machine learning, data quality is important because algorithms may be sensitive to noise and outliers. By measuring the degree of class overlapping and skewness in a dataset, \citet{Smith2013AnIL} proposes a generic definition of instance hardness representing how likely an example will be misclassified. \citet{Prudncio2015AnalysisOI} motivates from item response theory (IRT)~\citep{Embretson2000ItemRT, Ayala2008TheTA} to characterize instance hardness. \citet{Smith2011ImprovingCA} shows that properly removing hard examples in the dataset improves performance for a variety of learning algorithms including but not limited to Decision Tree and Support Vector Machine.

In deep learning regime, models with large capacity are typically more robust to outliers. Nevertheless, data examples can still exhibit diverse levels of difficulties. \citet{Arpit2017ACL} finds that data examples are not learned equally when injecting noisy data into training. \citet{Toneva2019AnES} shows that certain examples are forgotten frequently during training, which means that they can be first classified correctly then incorrectly. Model performance can be largely maintained when removing those least forgettable examples from training. \citet{Zhou2020CurriculumLB} proposes to dynamically estimate instance hardness during training and encourage the model to focus on those hard examples from a curriculum learning~\citep{Bengio2009CurriculumL} perspective, which can improve both the performance and efficiency for a wide range of datasets and neural architectures. More generally, under a self-paced learning~\citep{Kumar2010SelfPacedLF} framework, diverse methods have been proposed to mine hard examples on the fly~\citep{Chang2017ActiveBT}.

\subsection{Data profiling in robust learning}
In robust learning regime, the model is required to learn features that are robust to perturbations. Such task is generally more difficult, where the data examples are thus more likely to differentiate in terms of their behaviours in learning or contribution to the model performance. Plenty of works have analyzed the diverse behaviours of data during adversarial training, and proposed a variety of methods that treating the data examples differently. A detailed review has been made in Section~\ref{sect:related}, and Section~\ref{sect:tradeoff}, \ref{sect:robustoverfitting}, \ref{sect:gradientmask} that focus on existing problems in adversarial training specifically.

Here we mainly review works that investigate adversarial examples from a data perspective. \citet{Hendrycks2019NaturalAE} collects a set of unperturbed images, known as the ``natural adversarial examples'', that significantly degrades the performance of state-of-the-art image classifiers. \citet{Pestana2020AdversarialPP} observes that the adversarial perturbations concentrate in the Y-channel of the YCbCr space, which is believed to contain more shape and texture related information. \citet{Pestana2020DefensefriendlyII} reports the existence of a set of images that are particularly robust to adversarial perturbations. Including such data in validation significantly limits the reliability of the robustness evaluation.

% \subsection{Additional data: for furture work}
% robust self-training~\citep{Raghunathan2020UnderstandingAM}
\section{More experiment results}
% on the problems in adversarial training}
\label{sect:more-result}

\subsection{Data quality and robustness}
\label{append:perturbation-size-capacity}

\begin{figure*}[!ht]
\begin{subfigure}{0.48\textwidth}
% \begin{wrapfigure}{r}{0.55\linewidth} % [!ht]
% \vspace{-4mm}
  \centering
  \includegraphics[width=1.0\linewidth]{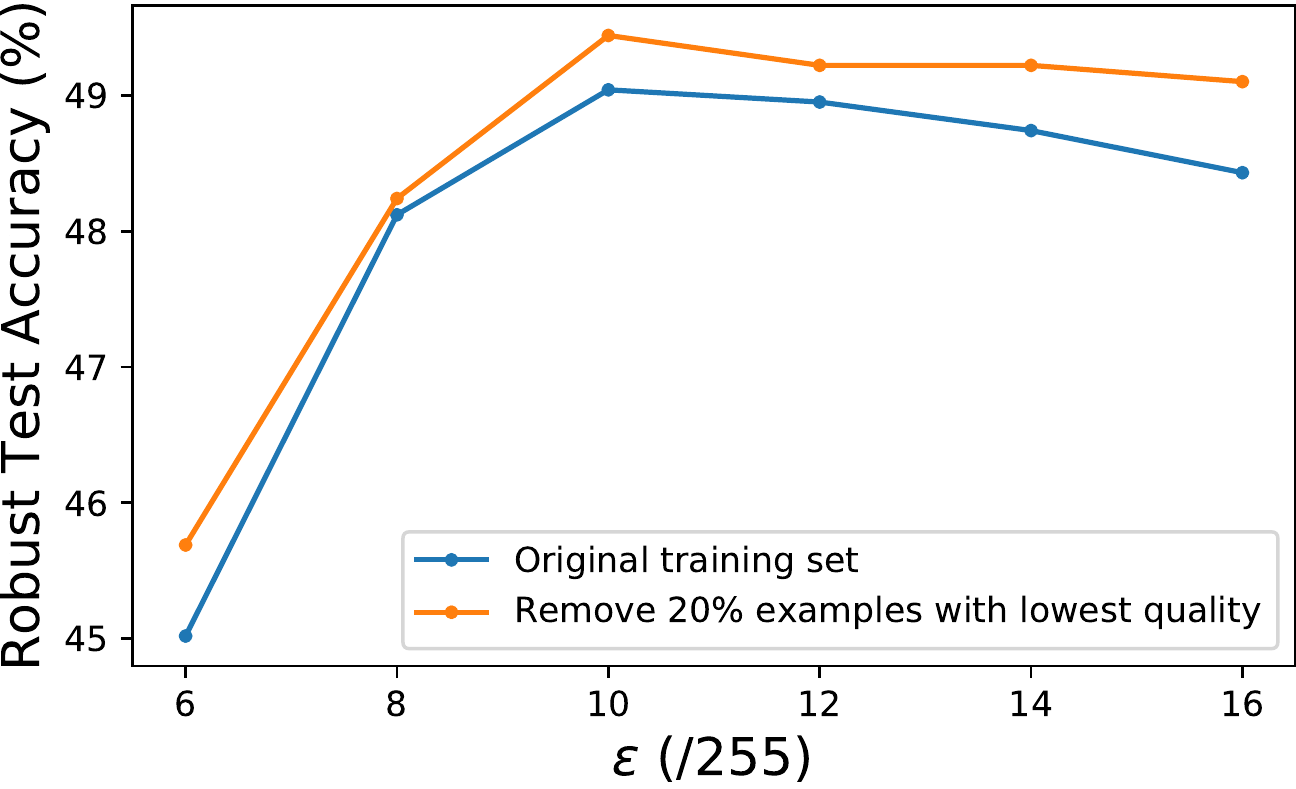}
  \caption{Removing a set of low-quality examples improves the robustness consistently for different training perturbation radii $\varepsilon$. The robustness improvement is more prominent for larger perturbation radii ($\ge 10/255$).
  % As the perturbation radius  varies from 6/255 to 16/255, and for both PGD and TRADES, the robustness obtained on the high-quality subset consistently outperforms that on the entire training set. 
  Here we remove $20\%$ examples with the lowest quality. We conduct adversarial training with PGD. The model is fixed as pre-activation ResNet-18.
  Other details of the experimental settings are mentioned in Appendix~\ref{sect:method-detail}.
  }
\label{fig:method-eps}
% \vspace{-2mm}
% \end{wrapfigure}
\end{subfigure}
\hfill
\begin{subfigure}{0.48\textwidth}
\centering
\includegraphics[width=1.0\textwidth]{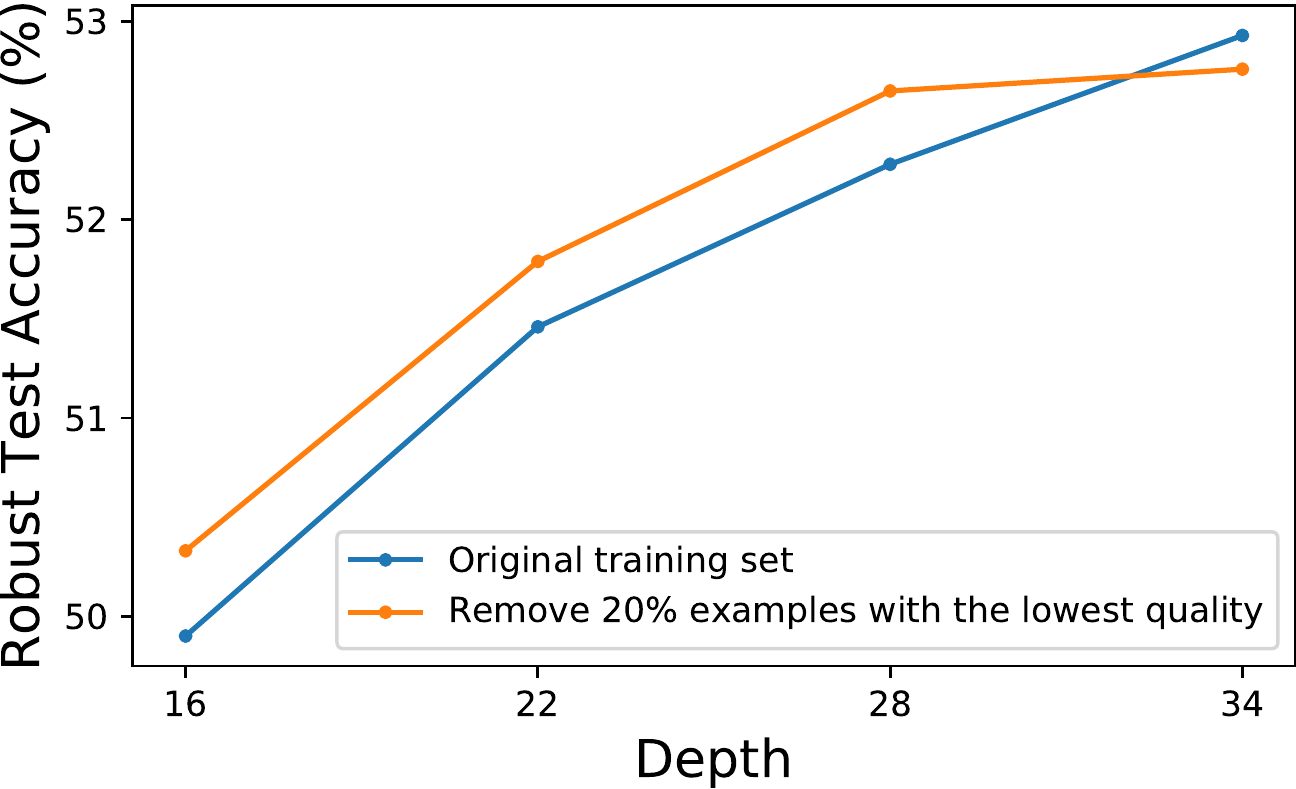}
\caption{The robustness improvement by removing a set of low-quality examples is more prominent for smaller models, and gradually vanishes as the model capacity increases. 
% However, removing less low-quality examples may still be beneficial to large models. 
Here we control the model capacity by modulating the depth of Wide ResNet, where the width is fixed as $10$. We experiment on PGD with the training perturbation radius yielding the best robustness ($12/255$). }
%   \centering
%   \includegraphics[width=1.0\textwidth]{figure/method-tradeoff.pdf}
%   \caption{The trade-off curve (robust test accuracy v.s. standard test accuracy) by varying the perturbation radius $\varepsilon$ on a basis of $2/255$, for different size of the high-quality subset $k$ used in training. $k=50$ indicates the training conducted on the entire training set. Other experimental settings are mentioned in Appendix~\ref{sect:method-detail}. Here $k$ amounts to a new hyperparameter that extends the trade-off to the robustness side. Stars denote the Pareto frontier. The axes are in log scale to show the highest accuracy and robustness clearly.}
% \label{fig:method-tradeoff}
\label{fig:method-capacity}
\end{subfigure}
\end{figure*}

\smallsection{Experiments on different models and perturbation sizes}
% \jingbo{Is this the place that you mentioned earlier to move to appendix? I feel this paragraph is off-the-topic.}
For larger perturbation radii and smaller models, removing a certain group of low-quality examples yields more robustness improvement as shown in Figure~\ref{fig:method-eps} and Figure~\ref{fig:method-capacity} respectively.
% Note that with problematic data being removed, TRADES can also benefit from a larger perturbation radius similar to PGD, which has been shown to be infeasible on the entire training set~\citep{Gowal2020UncoveringTL}. 
% As shown in Figure~\ref{fig:method-eps}, the improvement is more prominent when the perturbation radius is large, since the characteristic features of an image might be distorted more significantly by an adversary with larger perturbation.
% \jingbo{We don't have to stick with the same percentage number for the two figures. We just want to showcase that removing certain amount of data could even improve the robustness. So it's okay to have different ratios. Consider to keep only two lines per figure.}
Note that in Figure~\ref{fig:method-capacity}, the robustness improvement yielded by removing $20\%$ examples with the lowest quality diminishes as the model capacity increases. For considerably large models such as WRN-34-10 it will eventually hurts. % yet in which case removing less examples (E.g. $10\%$) will still yield robustness improvement.
These analyses match our intuition that adversarial examples produced by large perturbation are hard to learn; and smaller models are hard to learn adversarial robustness.

\subsection{Data quality and robust overestimation}
\label{sect:more-result-overestimation}

\smallsection{Experiments on more robustness evaluation methods}
% \jingbo{I think this paragraph could be compressed to one sentence and move the rest to appendix.}
To exclude the possibility that AutoAttack itself is sensitive to the data property due to its sophisticated design, we take CIFAR-10 as an example and select $20\%$ as a representative fraction of low-quality examples being removed where the overestimation gap shrinks significantly.
We further evaluate the robustness obtained by removing $20\%$ low-quality examples against a variety of black-box attacks including Square Attack, RayS and transfer attack from a separately trained surrogate model (See Appendix~\ref{sect:method-detail} for details). 
% To exclude the possibility that AutoAttack itself is sensitive to the data property due to its sophisticated design, we select a characteristic size of the friendly set ($4\times 10^4$) where the overestimation gap for PGD training starts to expand in Figure \ref{fig:overestimation-size} and further evaluate the robustness against black-box attacks including Square Attack~\citep{Andriushchenko2020SquareAA}, RayS~\citep{Chen2020RaySAR} and transfer attack from a separately trained surrogate model (See Appendix~\ref{sect:method-detail} for details). %  that is  on the original training set. 
We conduct repeated experiments ($5$ times) to reduce the statistical bias. 
We find that compared to the training set with $20\%$ low-quality examples removed, the robustness obtained on the entire training set can be significantly higher ($\sim 2\%$) when evaluated against PGD-10 and PGD-1000, i.e. PGD attack with up to 1000 iterations. However, the improvement is actually not clear if evaluated against Square Attack and RayS in that the average difference is less than the standard deviation, and may even be negative for AutoAttack and transfer attack.
% We find that when adding the most $10^4$ problematic data into the training set, the robustness against PGD-10 and PGD-1000~\footnote{} improves significantly ($\sim 2\%$), while the robustness against other attacks has no significant improvement for Square Attack and RayS (average change less than the standard deviation), and even degrades for AutoAttack and transfer attack.
In contrast, compared to the training set with $20\%$ randomly selected examples removed, the robustness obtained on the entire training set is consistently higher for all evaluation attacks.
% In contrast, when increasing the size of a randomly sampled dataset from $4\times 10^4$ to $5\times 10^4$ , the robustness against all attacks consistently improves ($\sim 1\%$).
This demonstrates that the low-quality data can cause difficulty for PGD attack to reliably evaluate the robustness, which cannot be avoided even with a substantially large number of iterations. This also implies PGD attack is likely to be suboptimal with fixed parameters~\citep{Mosbach2018LogitPM, Croce2020ReliableEO}.

% \subsection{Other neural architectures}

% \todo{Experiments to intro figure}
\section{More about data quality}

\subsection{Calculation of the data quality rank}
\label{sect:pl-calculation}

% \jingbo{we need a better name for this subsubsection. maybe a new notation? It's more like $\mathbb{E}[\rho(x)]$ and approximated by the average of 10 runs. Let's try to define the problematic likelihood by this expectation if you feel comfortable. And then, from the IoU and correlation analysis, we know it's quite stable. So 10 runs become a reasonable choice balancing the estimation accuracy and the running time cost.}

% \chengyu{If learning rate is sufficiently small and the training is converged, if there are sufficient number of experiments, our estimation should approach the true problematic likelihood.}

% \input{figScript/Identify-distribution}

To output a relatively accurate estimation of the data quality rank, we synthesize the results of multiple experiments. Specifically, we adversarially train a pre-activation ResNet-18 using PGD-10 on the entire CIFAR-10 training set for $160$ epochs with learning rate initialized as $0.1$ and decayed at epoch $80$ and $120$ with a factor of $10$. We repeat the training 10 times and average the data quality ranks calculated on each example, which yields a relatively stable estimation\footnote{Due to computational constraints, this estimation process only provides about $1600$ unique values for data quality ranks, which are not enough to differentiate all the $5\times10^4$ training examples in CIFAR-10. However, in above analyses, we at most partition the training set into $10$ subsets with different quality levels, which only results in about $0.6\%$ indistinguishable examples between any two adjacent subsets. One can seek more accurate estimation of data quality rank by incorporating the results from more experiments.}.
% Fig.~\ref{fig:problematic-distribution} shows the distribution of $\bar{\rho}(x)$ of the entire training set in CIFAR-10.

% ---------------------------------
\subsection{Distribution of learning stability}
\label{section:pl-distribution}
In Section \ref{section:problematic-method}, we estimate the data quality relatively by calculating the ranking of the examples based on learning stability. Directly using learning instability to measure the data quality might be troublesome as its magnitude varies greatly across different training settings (e.g. number of training epochs). % , training methods and neural architectures.

\begin{figure}[!ht]
  \centering
  \includegraphics[width=1.0\linewidth]{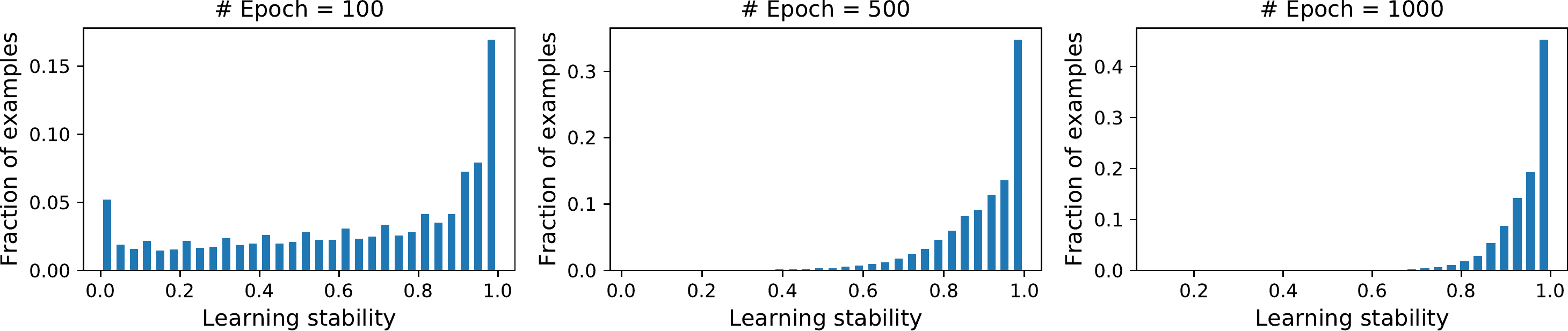}
  \caption{The distribution of learning stability given different training epochs.}
  \label{fig:distribution-stability-epoch}
\end{figure}

% \jingbo{drop the following two sentences or change it to ``refer to appendix'' style. It's like ``hey if you want to know its absolute values, refer to appendix''}
Figure~\ref{fig:distribution-stability-epoch} shows the distribution of the learning stability with different training epochs. 
One may observe that, the overall magnitude of the learning stability varies greatly as the training setting changes, which is reasonable since the model complexity hinges on the training settings and all training examples will eventually be overfitted given sufficient training epochs. 

% We also append the distribution of the learning stability on different datasets for reference. One can expect that CIFAR-100 and Tiny-ImageNet are significantly more difficult since less num
% % Larger models can learn more problematic data thus the distribution is shifted to the left.

\section{More methods to measure the data quality}
\label{sect:broad}

In this section, we show that it is possible to estimate the data quality motivated from multiple measurements including prediction probability, minimum perturbation and learning order. Each measurement is itself consistent across different training settings, and is also correlated with the data quality rank estimated in Section~\ref{sect:pl-estimation}. 
% This further demonstrates the existence of the problematic score that is intrinsic to the data and profoundly influences the adversarial training. 
We also briefly review existing adversarial training methods leveraging these measurements to customize the inner maximization or outer minimization for individual examples in their designs, which implies that their improved performance is potentially due to treating examples of different data quality differently.

% \subsection{Frequency of forgotten events}
% \label{sect:broad-forgotten}

% \todo{A little bit of theory}

\subsection{Prediction probability}
\label{sect:broad-probability}

Based on the standard adversarial training~\citep{Madry2018TowardsDL}, multiple variants pivot on the utilization of soft output in the adversarial loss function. Here we specifically refer the soft output as the either the output before the softmax function, namely logit, or that after the softmax function, namely prediction probability. Adversarial logit pairing (ALP)~\citep{Kannan2018AdversarialLP} is a method explicitly penalizing the difference between the logits from an clean example and its adversarial counterpart on top of the standard cross-entropy loss for adversarial training. BGAT~\citep{Cui2020LearnableBG} instead collects the logits of clean examples from an auxiliary clean model. GAT~\citep{Sriramanan2020GuidedAA} adopts a similar loss function penalizing probability difference instead of logit difference.
VAT~\citep{Miyato2019VirtualAT} and TRADES~\citep{Zhang2019TheoreticallyPT} propose a loss function matching the probability from an adversarial example with its clean counterpart, instead of the true label.
Self-adaptive training~\citep{Huang2020SelfAdaptiveTB} and MART~\citep{Wang2020ImprovingAR} use the probabilities from clean examples to weight the examples in the loss function, although the former focuses more on examples with high probabilities while the latter focuses more on examples with low probabilities.

% LLR and Curvature regularization? Seems all same. % not quite, they are first-order approximation, mainly focus on the regularization of gradients

% \chengyu{knowledge distillation? also use soft output, but from another larger model}

Prediction probability is also related to whether an example is correctly classified or not. Low probability of the true label indicates an example is likely to be misclassified, which might be troublesome in adversarial training because adversarial examples of misclassfied examples are ``undefined''~\citep{Wang2020ImprovingAR}. Several methods are thus motivated to treat correctly classified and misclassified examples differently. MMA~\citep{Ding2020MaxMarginA} employs adverarial training only on correctly classified examples, leaving misclassified examples to standard training. MART~\citep{Wang2020ImprovingAR} introduces a term associated with the probability of the true label in the loss function to encourage the learning on misclassified examples. A summary of the loss functions employed in these methods can be found in Table 1 by \citet{Wang2020ImprovingAR}.

% Probability estimated by multiple models also indicates how likely an example will be correctly classified (\cite{Ding2020MaxMarginA, Wang2020ImprovingAR}). 

% \begin{table*}[!h]
%   \small
%   \caption{\note{introduce the notation}}
%   \label{table:soft-methods}
%   \centering
% %   \small
%   \begin{tabular}{lc}
%     \toprule
%     Method & Loss Function \\
%     \midrule
%     Standard & $\text{CE}(p(x+\delta), y)$\\
%     \bottomrule
%     \footnotesize{$^*$ The best hyper-parameter obtained.}\\
%   \end{tabular}
% \end{table*}

% Margin encouraged by probability~\citep{Madry2018TowardsDL, Ding2020MaxMarginA, Wang2020ImprovingAR}. 

% High prediction probability indicates an example can sustain larger perturbation with fixed model Lipschitz. (any work? this is more like instance margin)

Recently, adversarial training with additional data becomes increasingly popular. Prediction probability is used to identify high-quality data that is more relevant to the original distribution~\citep{Gowal2020UncoveringTL}.

% And also in that test set analysis paper.

\begin{figure}[!ht]
\centering
\begin{subfigure}{0.3\textwidth}
  \centering
  \includegraphics[width=0.95\linewidth]{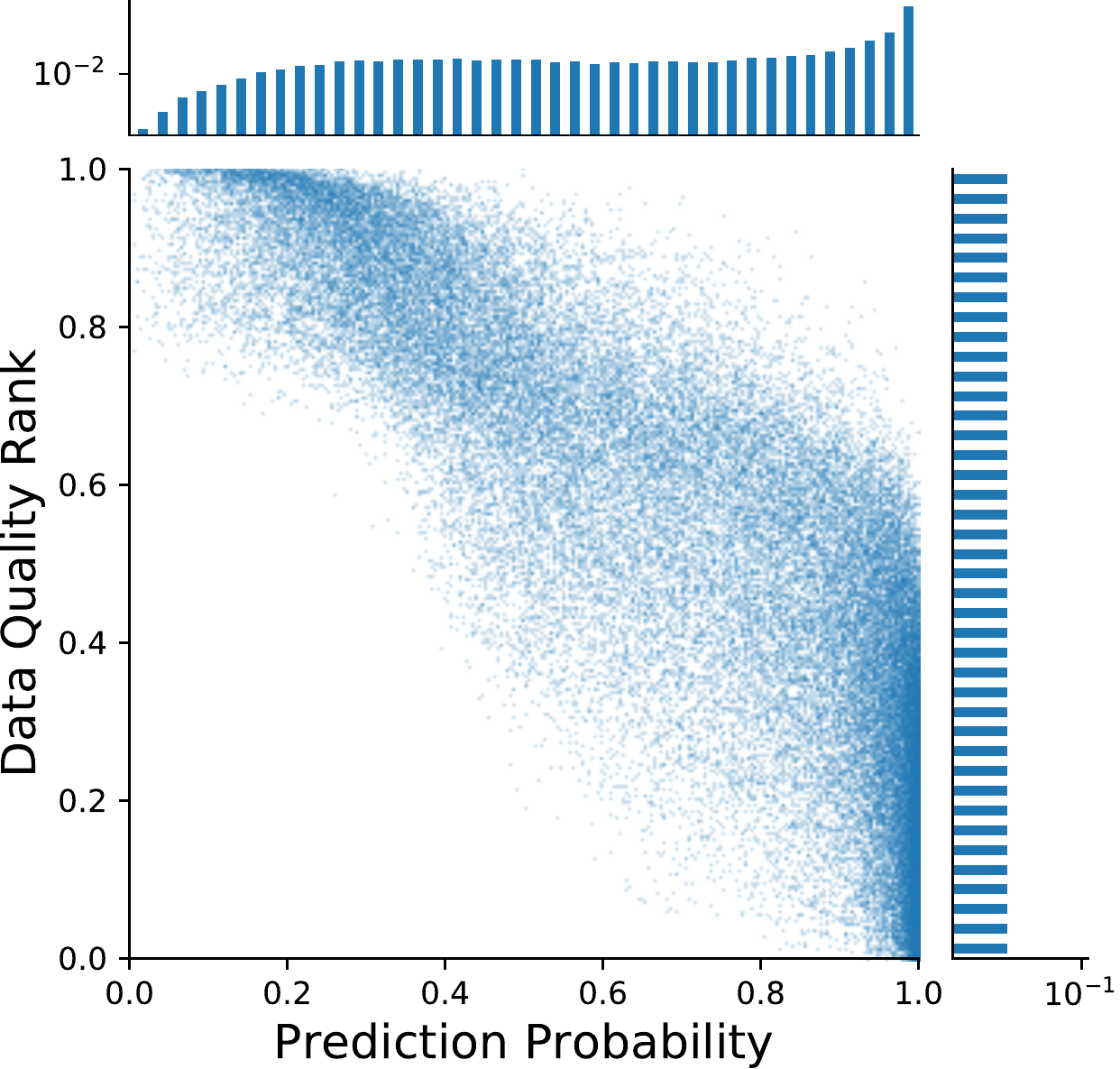}
  \caption{Prediction probability}
  \label{fig:corr-conf-pro}
\end{subfigure}%
\begin{subfigure}{0.3\textwidth}
  \centering
  \includegraphics[width=0.95\linewidth]{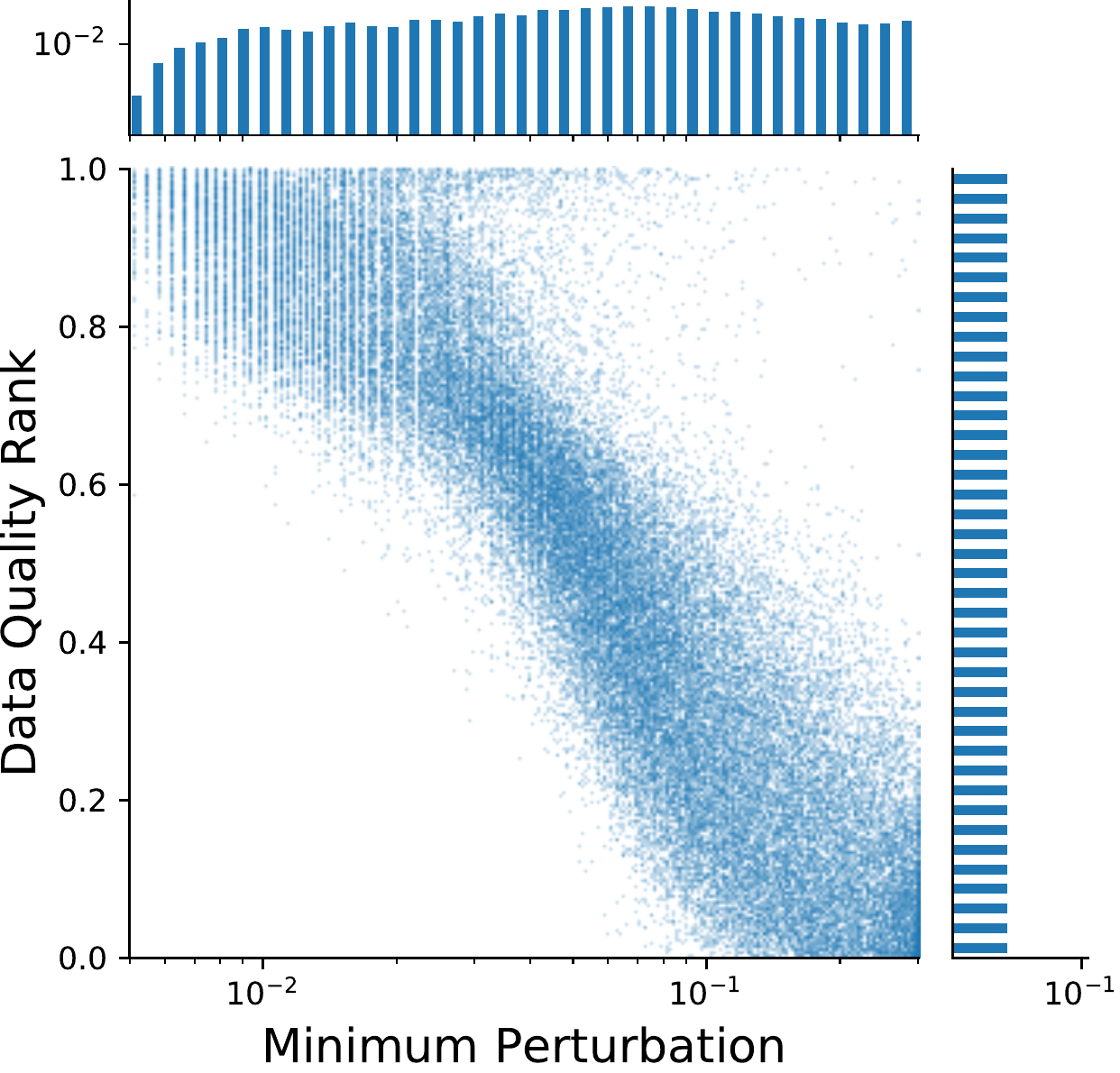}
  \caption{Minimum perturbation}
  \label{fig:corr-minperturb-prob}
\end{subfigure}
\begin{subfigure}{0.3\textwidth}
  \centering
  \includegraphics[width=0.95\linewidth]{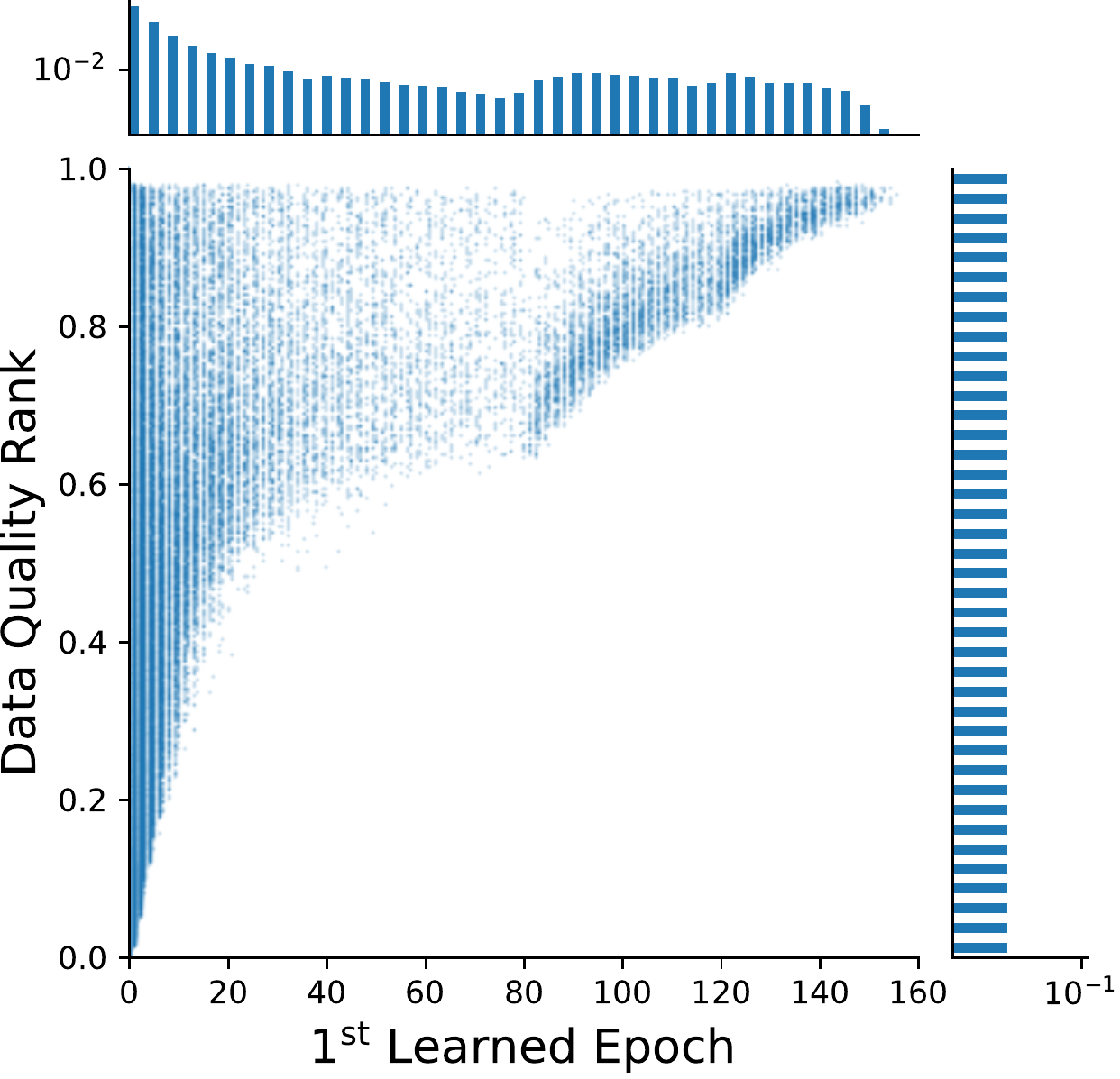}
  \caption{Learning order}
  \label{fig:corr-learningorder-prob}
\end{subfigure}
\caption{Correlation between data quality rank and other measurements}
\label{fig:extra-corr}
\end{figure}

Here we show that the prediction probability of an example is correlated with its data quality rank. These sophisticated methods are thus likely to achieve robustness gain by treating examples with different quality levels differently. We specifically refer the \emph{prediction probability} as the probability corresponding to the true label from a clean input, and use the best model in terms of robustness throughout training to estimate it. For every example, we average the probabilities obtained by the same $10$ experiments introduced in Section~\ref{sect:pl-calculation}. As shown in Figure~\ref{fig:corr-conf-pro}, prediction probability is inversely correlated with data quality rank. A high-quality example is inclined to be correctly classified by the model with high probability. We do not estimate the data quality based on whether an example will be misclassified or not since the misclassification rate varies greatly across different training settings. 

% \chengyu{This section leads to future works: 1. Why do some methods work (not work). What's the connection to problematic rank. 2. Are misclassified examples really helpful in adversarial training.}

% \note{Show to achieve the optimal robustness we need part of the examples that are likely to be misclassified. Simply put the result on the original binary friendly subset, show it is not as good as 40k subset.} - maybe future work

% =============================================
\subsection{Minimum perturbation} %  / Distance to boundary / Attackability}
\label{sect:broad-minimum-perturbation}
% It has been widely noticed that the data are not born equal in adversarial training. 
Standard adversarial training often sets a perturbation radius universal to all training examples. However, it has been widely noticed that individual examples may have different levels of robustness against adversarial attacks. % due to different levels of ambiguity. 
% It might not be possible to find an imperceptible perturbation level universal to all examples in adversarial training.
% It might not be possible to find a perturbation radius universal to all examples in adversarial training.
It might be helpful to customize the perturbation for each example during adversarial training.
MMA~\citep{Ding2020MaxMarginA} proposes a method to estimate proper individual perturbation for each example based on its distance to the decision boundary. The perturbation is determined by a line search along the perturbation direction initialized by a norm-constrained PGD attack. In a similar vein, IAAT~\citep{Balaji2019InstanceAA} performs a dynamic update of individual perturbation throughout the training. CAT~\citep{Cheng2020CATCA} further incorporates individual label smoothing based on the estimated perturbation level. Instead of customizing individual perturbation radius for each example, FAT~\citep{Zhang2020AttacksWD} customizes the number of attack iterations for each example, such that the perturbation is just enough to fool the model. GAIRAT~\citep{Zhang2020GeometryawareIA} further adopts a weighted loss function based on such individual attack iterations to focus more on those examples far from the decision boundary. Nevertheless, none of these works detailedly analyzes the impact of such individual perturbation on the adversarial training.

% Or simply the label-flipping distance following the gradient direction.

Here, we show that profiling the data based on individual perturbation radius leads to similar result as that motivating from learning stability introduced in Section~\ref{sect:pl-estimation}. We denote the \emph{minimum perturbation} of an example as the smallest perturbation radius required to change a model's prediction on it. Ideally, the minimum perturbation of an example amounts to its minimum distance to the decision boundary. We use an untargeted attack based on I-FGSM (Iterative Fast Gradient Sign Method) \citep{Goodfellow2015ExplainingAH, Kurakin2017AdversarialML} with step size $1/255$ based on the implementation in Adversarial Robustness Toolbox (ART) \citep{art2018}. We empirically found the step size $1/255$ is often small enough to ensure a converged minimum perturbation. One can also employ other attacks including but not limited to DeepFool~\citep{MoosaviDezfooli2016DeepFoolAS}, CW~\citep{Carlini2017TowardsET}, EAD~\citep{Chen2018EADEA} and FAB~\citep{Croce2020MinimallyDA} in estimation. % See introduction of fast adaptive boundary attack for more related work % \cite{curvature}? cannot remember what is this.
But note that all these methods cannot obtain the true minimum perturbation, but only the upper bound of it~\citep{Weng2018EvaluatingTR}. Estimation of the minimum perturbation through optimization is known as a NP-hard problem~\citep{Katz2017ReluplexAE}.

In Figure~\ref{fig:corr-minperturb-prob}, we use the best model in terms of robustness obtained in training to estimate the minimum perturbation, and average the results obtained by the same $10$ experiments mentioned in Section~\ref{sect:pl-calculation}. One can find that the minimum perturbation is inversely correlated with the data quality rank, which means that a low-quality example is more likely to reside near the decision boundary. This suggests the sophisticated methods mentioned above are likely to treat examples with different qualities differently. It also suggests that examples with different amounts of minimum perturbation will influence the adversarial training differently in terms of the contributions to robustness and aforementioned problems.

% % =============================================
% \subsection{Inter-class distance}
% \todo{maybe neglect}
% This is not exactly an identification method, but a verification method. Because it is mentioned in `on the sensitivity`
% % \input{figScript/Extra-iterations}

% =============================================
\subsection{Learning order}
\label{sect:learning-order}

% learning stability is the unranked problematic likelihood, there is a two peak might be interesting.

Learning order refers to the phenomenon that Deep Neural Networks (DNNs) learn the examples in a similar order, which is shared by different random initializations and neural architectures. Such phenomenon is observed widely in standard training and training with noisy inputs and labels~\citep{Arpit2017ACL, Li2020GradientDW}. It is demonstrated that the learning order originates from the coupling between DNNs and benchmark datasets~\citep{Hacohen2019LetsAT}, since DNNs learn synthetic datasets without a specific order and classifiers other than DNNs such as AdaBoost~\citep{Freund1995ADG} learn benchmark datasets without a specific order.

% For adversarial training (learned == adversarially learned):

% Learning order \cite{Hacohen2019LetsAT} <- just 1st learned (simpler - tp-agreement needs too many experiments)

% forgetting events (forgettable and unforgettable examples - problematic examples are more forgettable)\cite{Toneva2019AnES} <- Can use 50kx160 table to calculate

% memorization (Average probability after 1st epoch) (easy examples learned in the 1st epoch - problematic data are harder.)\cite{Arpit2017ACL}
% \cite{} <- Need the model at the 1st epoch. Should be quick.

% in noisy label regime, first learn clean label then noisy label (\cite{Li2020GradientDW})

% Learning order is true for both clean and adversarial learning setting.

We show that learning order still exists in adversarial training. We denote the \emph{$1^\text{st}$ learned epoch} as the first epoch when a training example is classified correctly under adversarial attack. We show there is a correlation between the $1^\text{st}$ learned epochs of an example across different training settings (see Section~\ref{sect:broad-correlation}). Furthermore, we show that learning order is correlated with data quality rank. In Figure~\ref{fig:corr-learningorder-prob}, we average the $1^\text{st}$ learned epochs of an example obtained by the same experiments as mentioned in Section~\ref{sect:pl-calculation}. One can find the $1^\text{st}$ learned epoch is positively correlated with its quality rank, which means low-quality examples are likely to be learned late during training. 

Note that if we pick the best epoch\footnote{The epoch when the best model in terms of robustness is obtained} as a boundary and partition the training examples based on their $1^\text{st}$ learned epochs, the resulting two subsets correspond to exactly the correctly classified and misclassified examples. This implies that those adversarial training methods treating examples differently based on whether they are correctly classified or not, as mentioned in Section~\ref{sect:broad-probability}, are likely to be a special case of treating examples differently based on their quality, from yet another perspective.

\subsection{Consistency of the motivation}
\label{sect:broad-correlation}

\begin{figure}[!ht]
    \centering
    \begin{subfigure}{0.33\textwidth}
      \centering
      \includegraphics[width=0.95\linewidth]{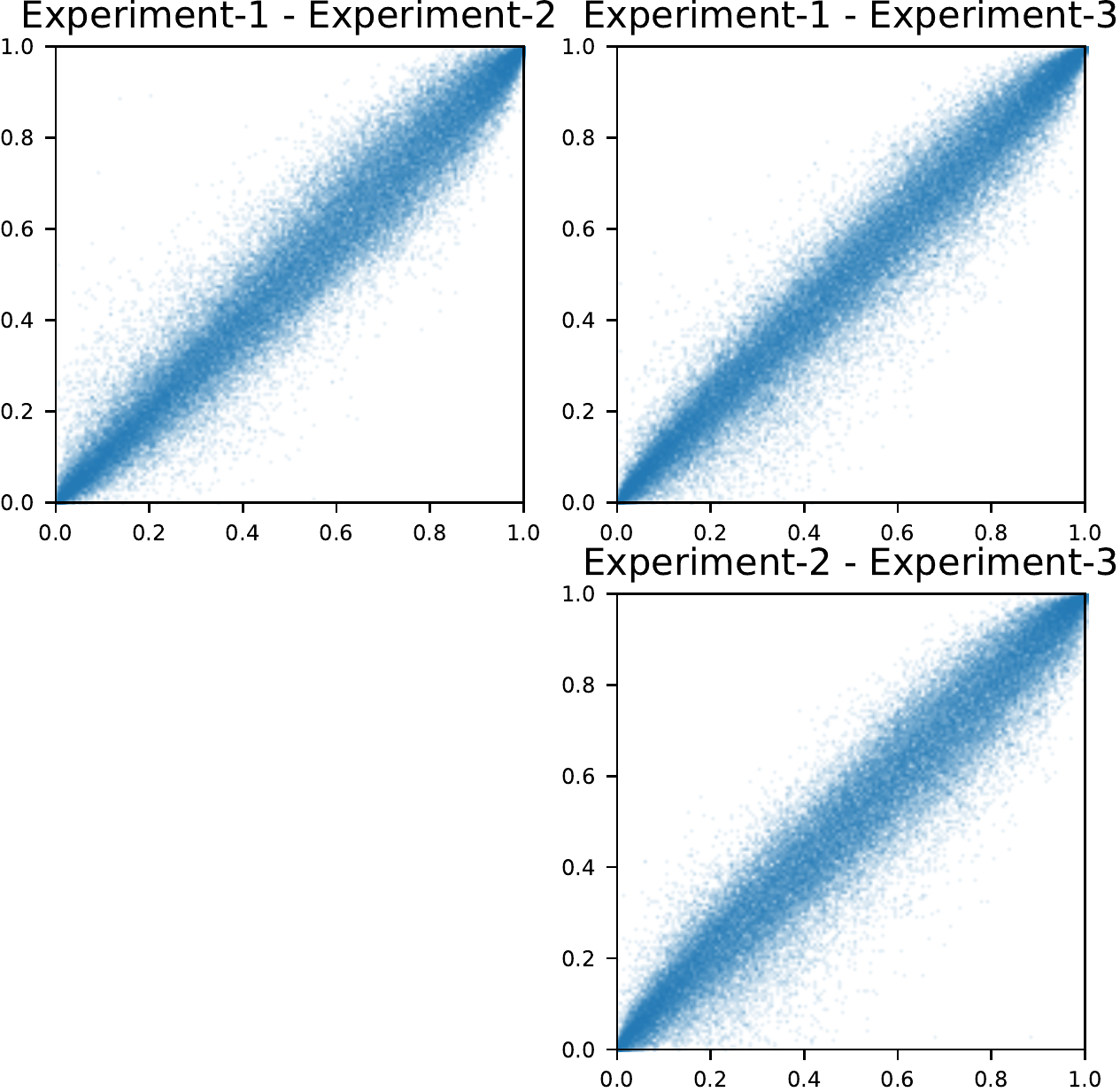}
      \caption{Random initializations}
      \label{fig:broad-corr-proba-experiments}
    \end{subfigure}%
    \begin{subfigure}{0.33\textwidth}
      \centering
      \includegraphics[width=0.95\linewidth]{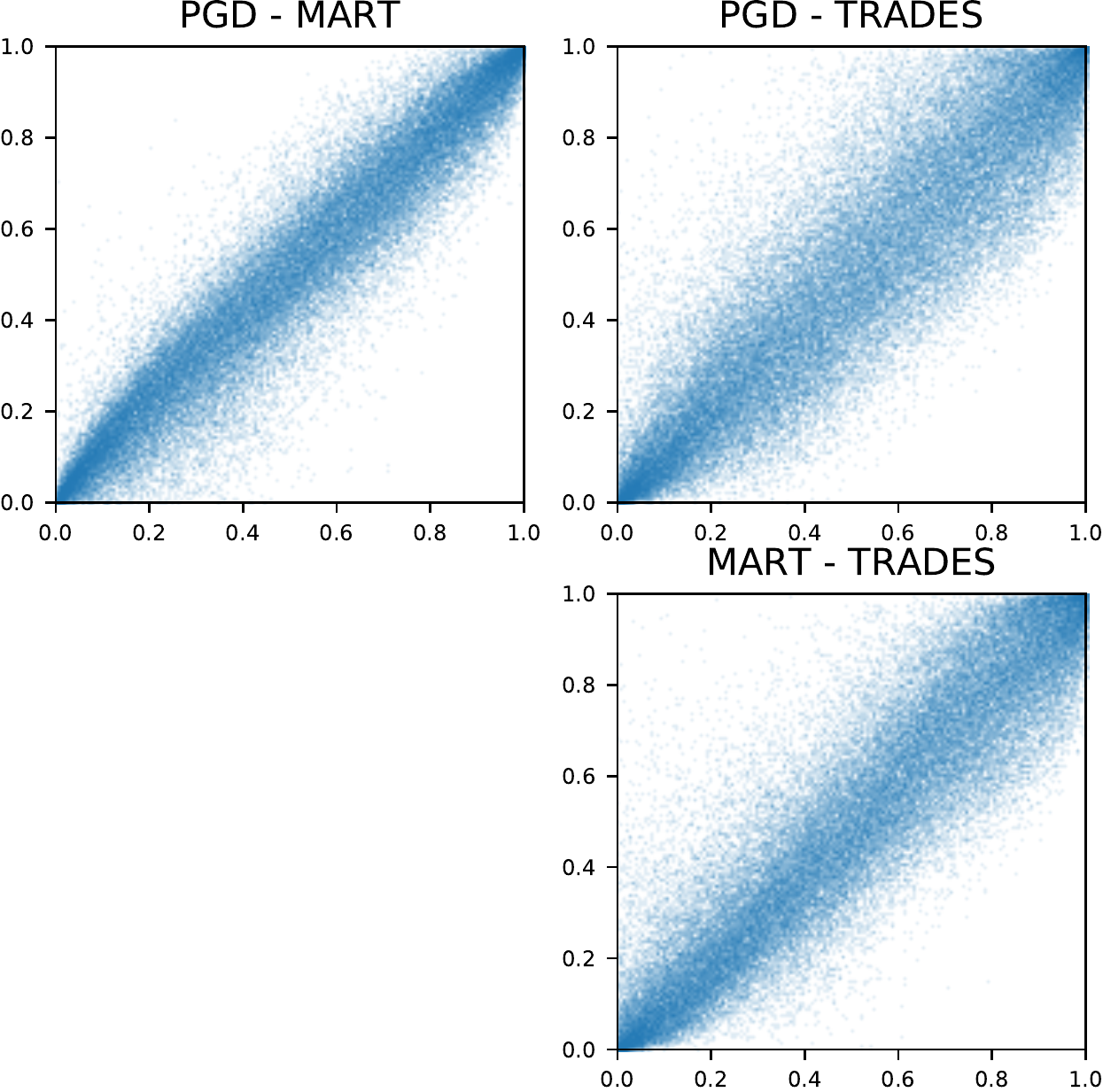}
      \caption{Different methods}
      \label{fig:broad-corr-proba-methods}
    \end{subfigure}
    \begin{subfigure}{0.33\textwidth}
      \centering
      \includegraphics[width=0.95\linewidth]{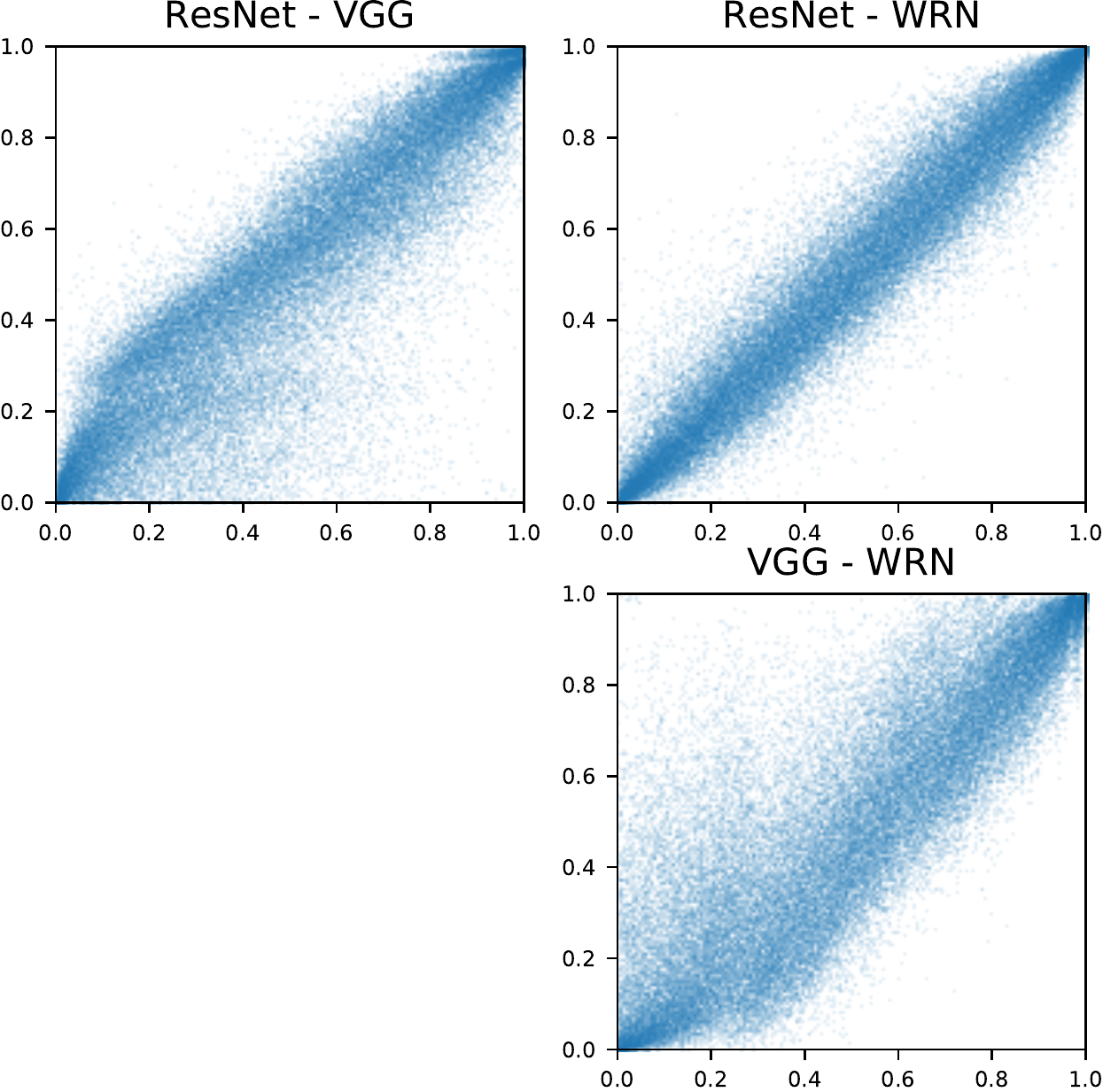}
      \caption{Different models}
      \label{fig:broad-corr-proba-models}
    \end{subfigure}
    \caption{Scatter plots of the quality ranks of training examples based on prediction probabilities obtained by different training settings. The prediction probability of an example is consistent across random initializations, different methods and models.}
    \label{fig:broad-corr-proba}
\end{figure}

\begin{figure}[!ht]
    \centering
    \begin{subfigure}{0.33\textwidth}
      \centering
      \includegraphics[width=0.95\linewidth]{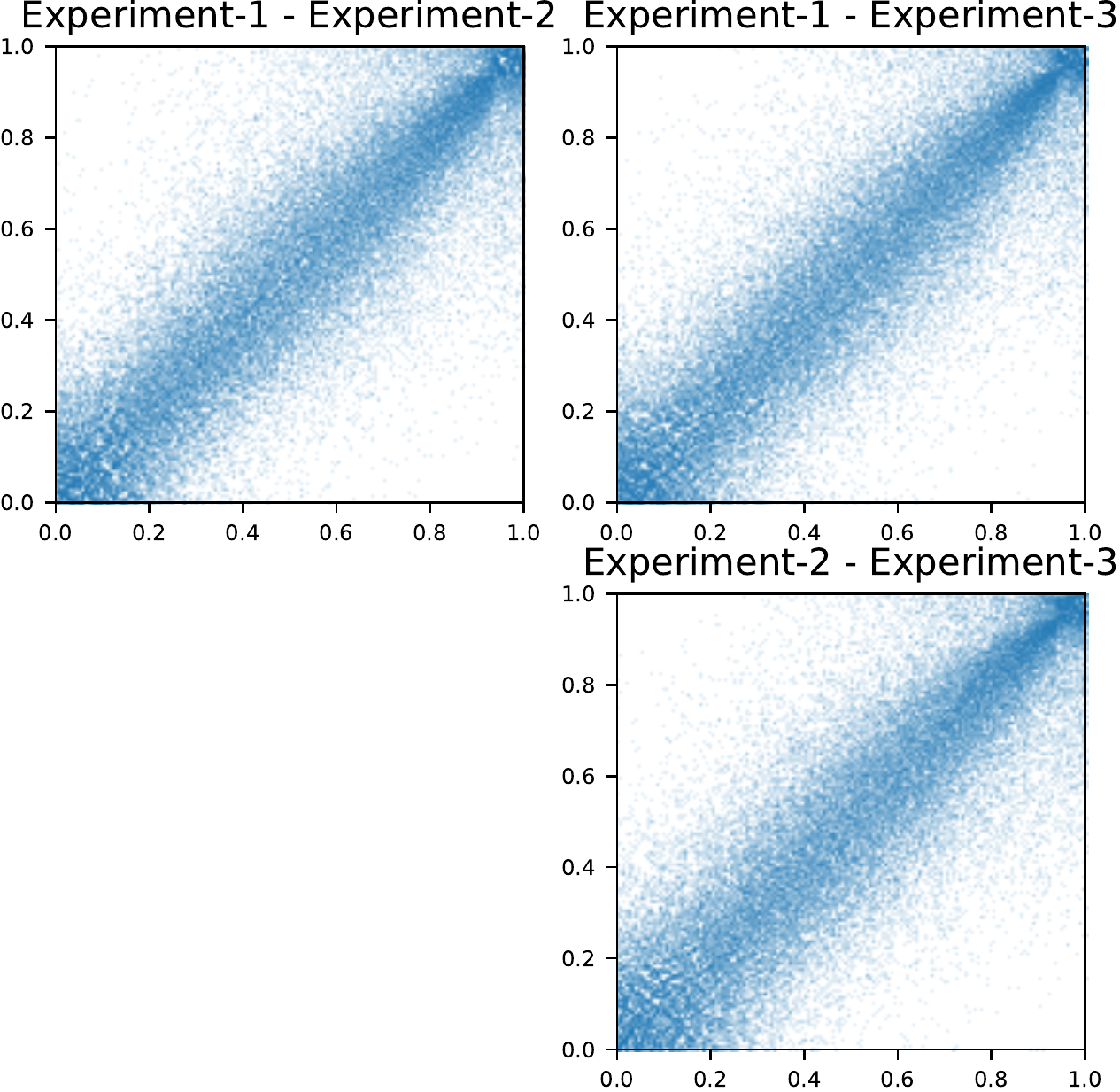}
      \caption{Random initializations}
      \label{fig:broad-corr-mindist-experiments}
    \end{subfigure}%
    \begin{subfigure}{0.33\textwidth}
      \centering
      \includegraphics[width=0.95\linewidth]{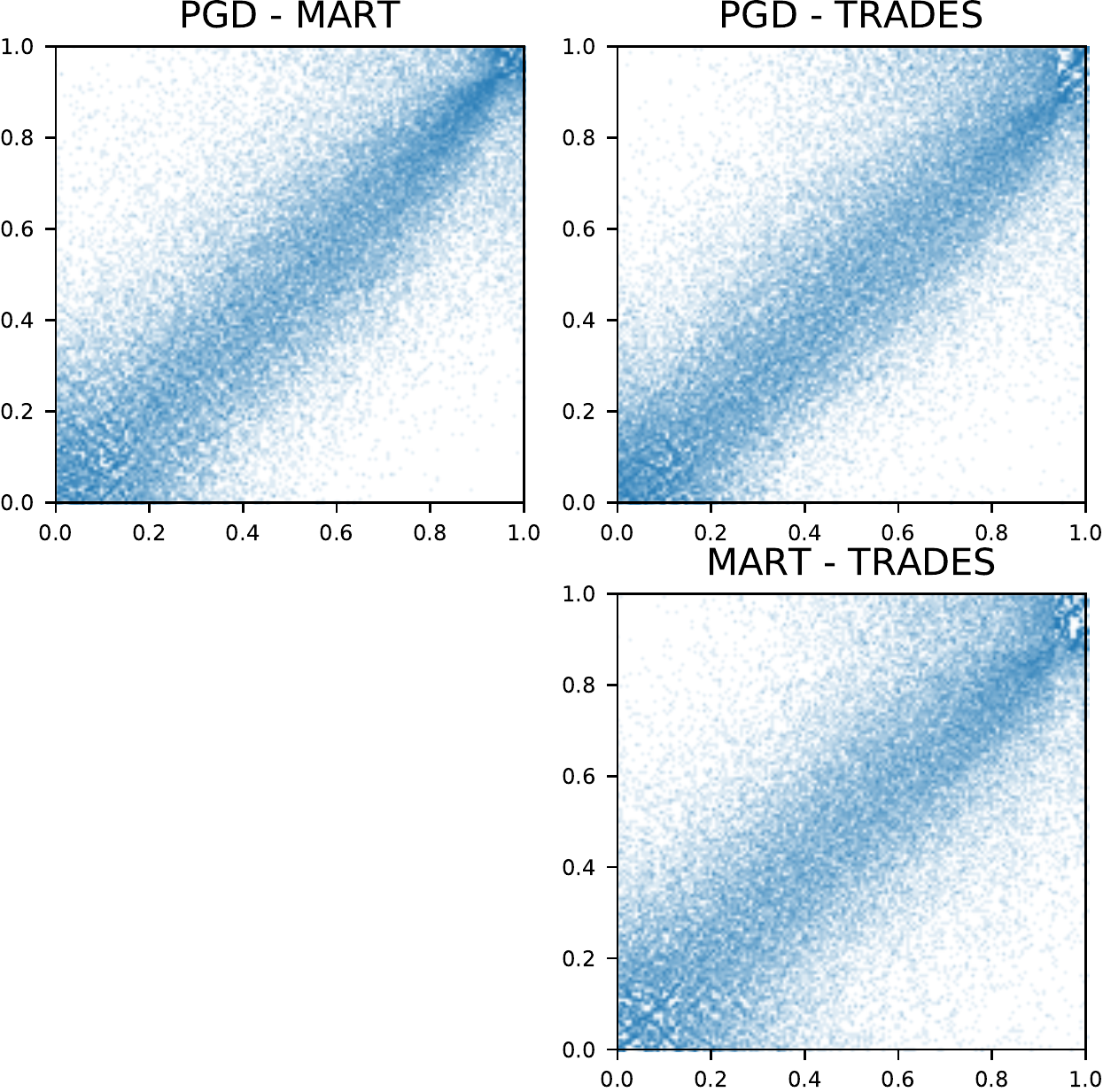}
      \caption{Different methods}
      \label{fig:broad-corr-mindist-methods}
    \end{subfigure}
    \begin{subfigure}{0.33\textwidth}
      \centering
      \includegraphics[width=0.95\linewidth]{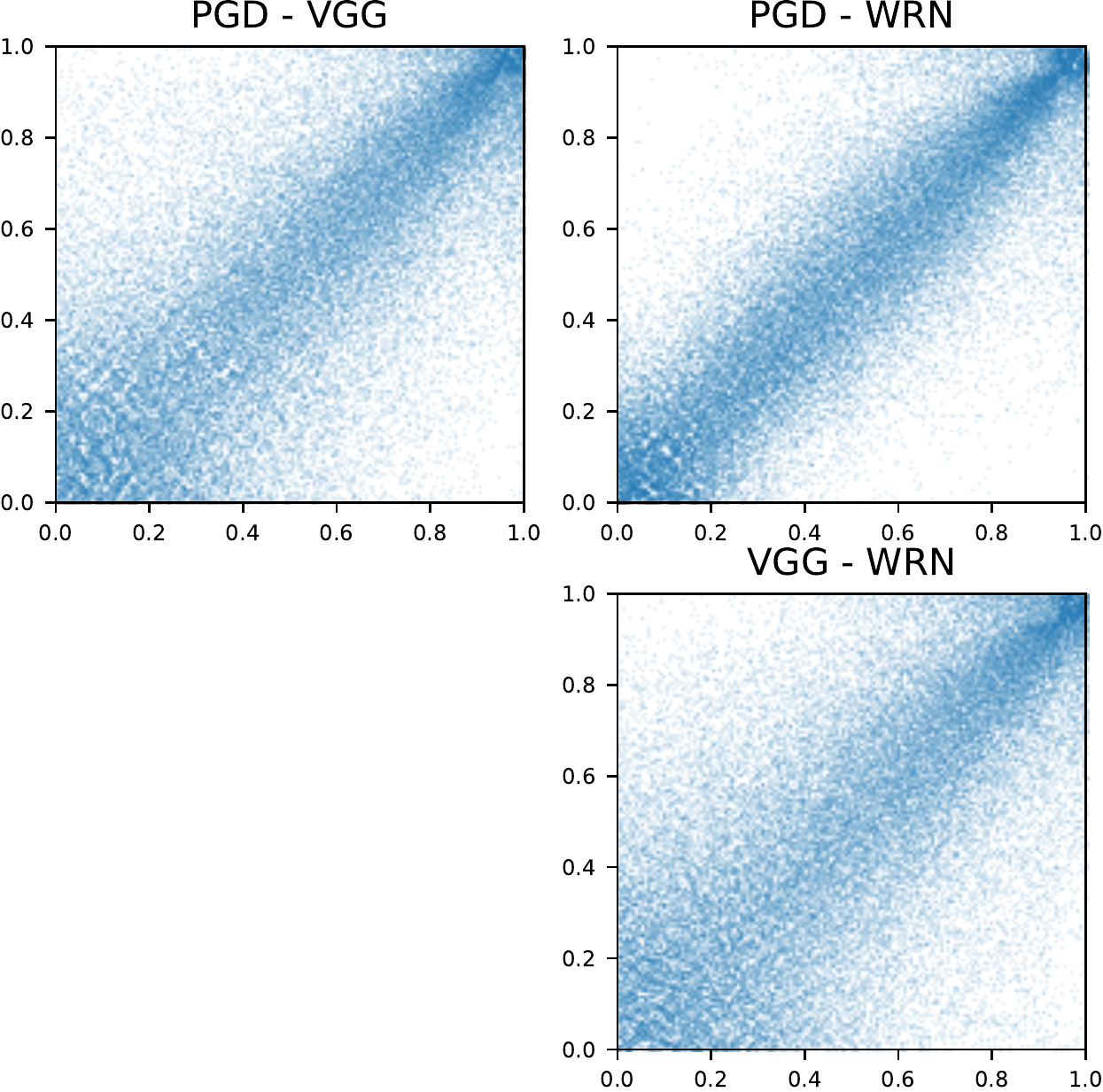}
      \caption{Different models}
      \label{fig:broad-corr-mindist-models}
    \end{subfigure}
    \caption{Scatter plots of the quality ranks of training examples based on minimum perturbations obtained by different training settings. The minimum perturbation of an example is consistent across random initializations, different methods and models.}
    \label{fig:broad-corr-mindist}
\end{figure}

\begin{figure}[!ht]
    \centering
    \begin{subfigure}{0.33\textwidth}
      \centering
      \includegraphics[width=0.95\linewidth]{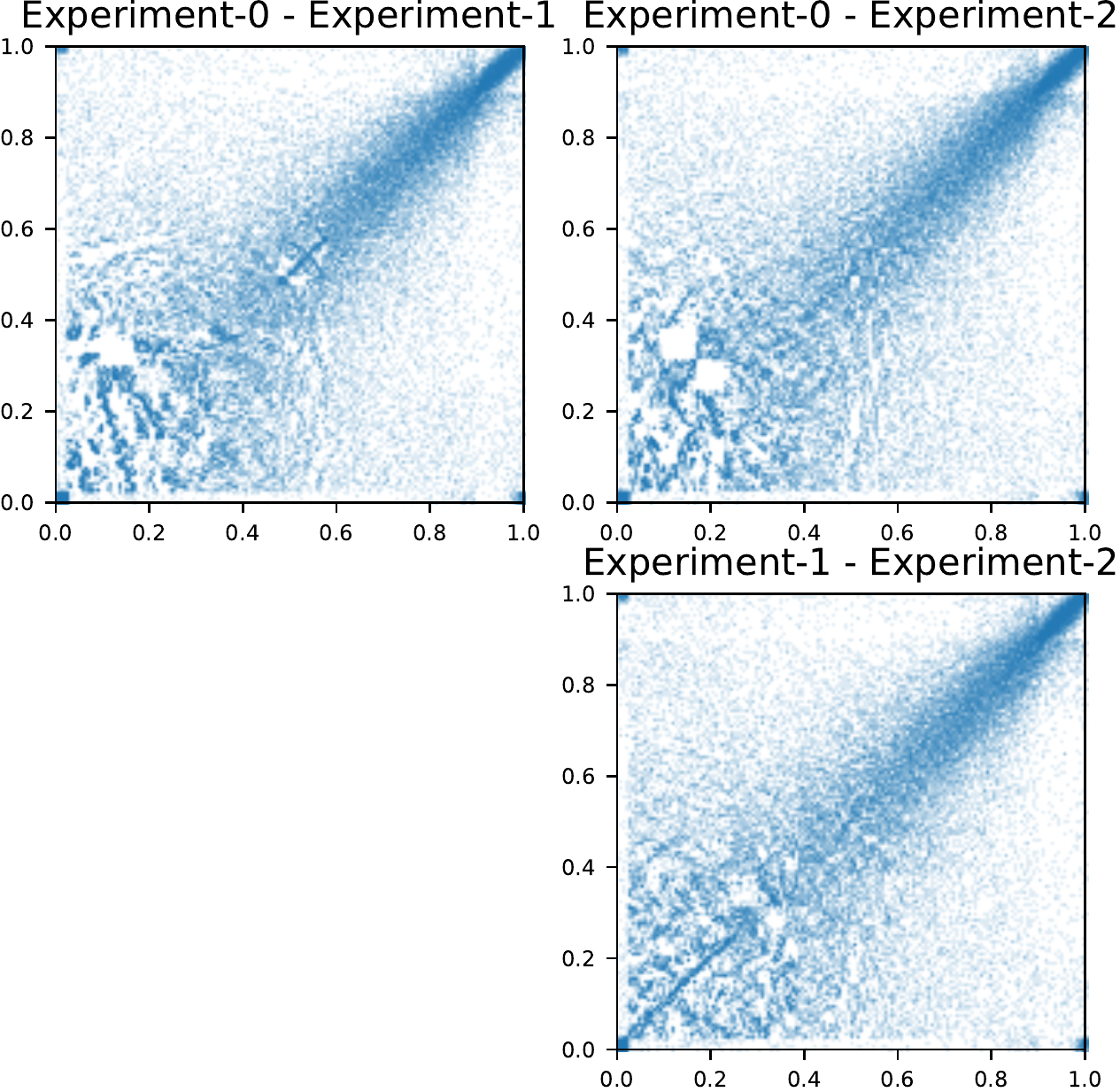}
      \caption{Random initializations}
      \label{fig:broad-corr-learningorder-experiments}
    \end{subfigure}%
    \begin{subfigure}{0.33\textwidth}
      \centering
      \includegraphics[width=0.95\linewidth]{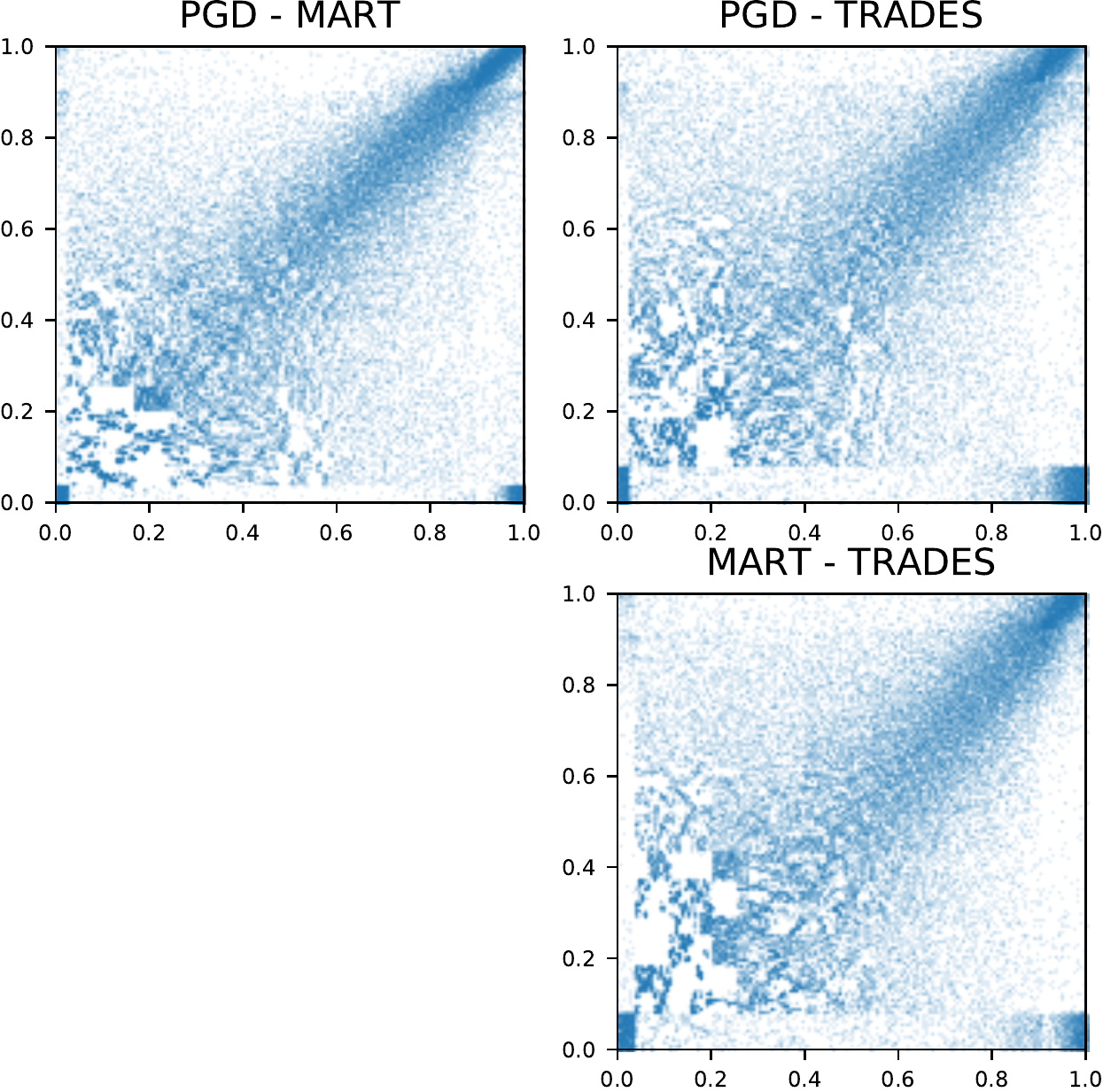}
      \caption{Different methods}
      \label{fig:broad-corr-learningorder-methods}
    \end{subfigure}
    \begin{subfigure}{0.33\textwidth}
      \centering
      \includegraphics[width=0.95\linewidth]{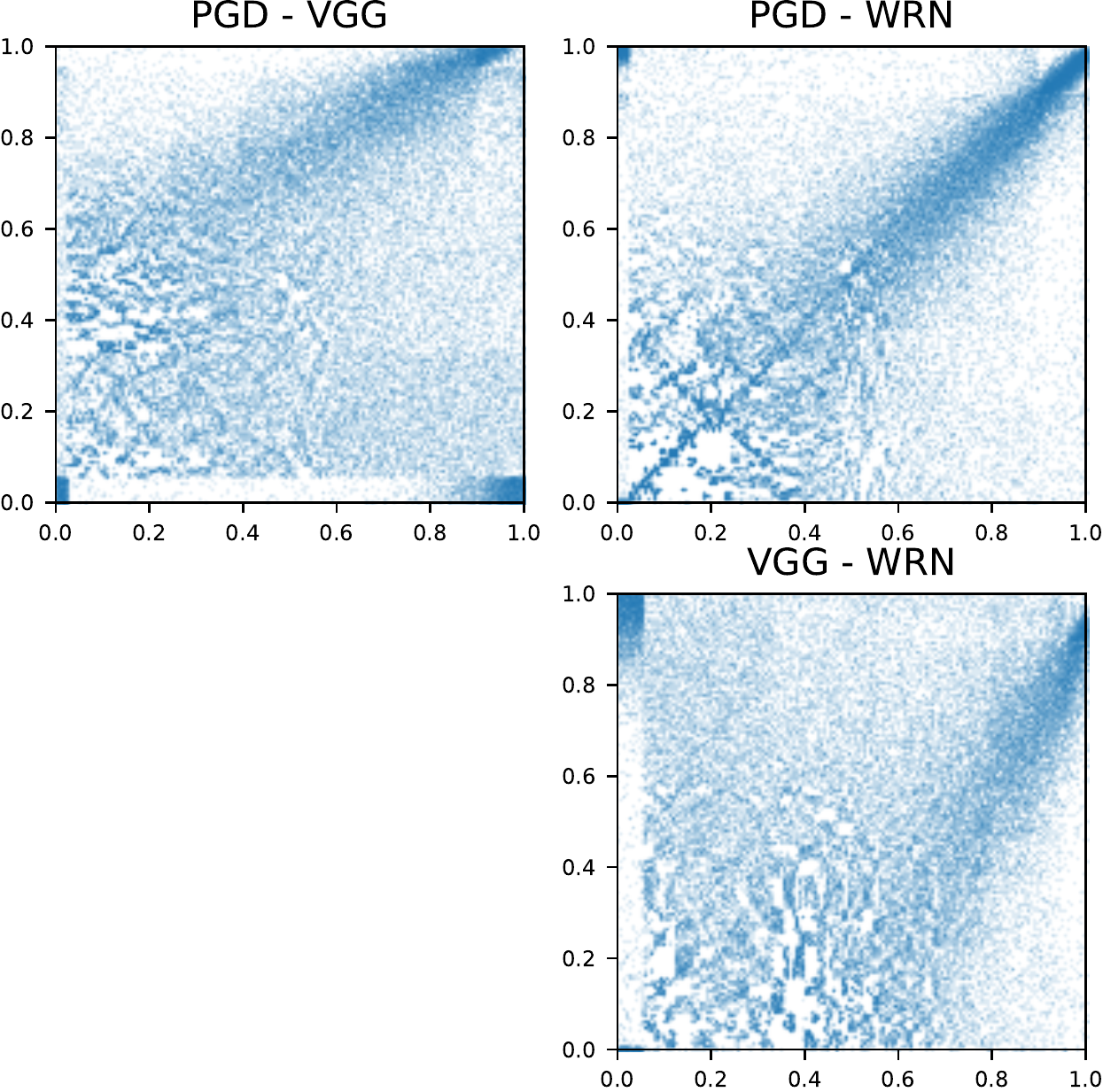}
      \caption{Different models}
      \label{fig:broad-corr-learningorder-models}
    \end{subfigure}
    \caption{Scatter plots of the quality ranks of training examples based on $1^\text{st}$ learned epochs obtained by different training settings. The $1^\text{st}$ learned epoch refers to the first epoch when an example is classified correctly under adversarial attack, which is consistent across random initializations, different methods and models.}
    \label{fig:broad-corr-learningorder}
\end{figure}

We show that each motivation mentioned above is itself consistent across different training settings. Therefore it is possible to estimate the data quality rank similarly from each motivation. In Figure~\ref{fig:broad-corr-proba}, ~\ref{fig:broad-corr-mindist}, ~\ref{fig:broad-corr-learningorder}, we use the same experiments mentioned in Figure~\ref{fig:agreement-corr} to show that the prediction probability, minimum perturbation and learning order are all consistent across random initializations, different training methods, and neural architectures. Nevertheless, one may find that these estimations are relatively less consistent compared to the data quality rank estimated from learning stability, which is the major reason that we estimate the data quality rank from learning stability in the main text.
% \chengyu{Also mention misclassified examples are not flexible (either 0 or 1). So we didn't use misclassified examples to identify the problematic examples either.}

% Empirically learning stability is the most consistent across initializations, methods and models.

% ---------------------------------
% ---------------------------------
\section{Limitations of removing low-quality data}
\label{sect:performance-tradeoff}

% ---------------------------------
% \subsection{Performance Trade-off induced by the amount of problematic data}

% \note{Change $k$ to some other variable. $N_r$? The number of problematic examples being removed}

We note that removing low-quality training data cannot be advocated as a completing method. Although removing low-quality data can benefit the robustness in certain scenarios, it will inevitably impair the standard accuracy, since those hard examples are of high-quality to standard training. As shown in Table~\ref{table:remove-hurt}, removing the $20\%$ examples with the lowest quality can improve the best robustness consistently for different training methods and perturbation radii, but will also diminish the standard accuracy for most settings (except PGD training with perturbation radius $12/255$).

% One may note that when training on the high-quality subset only, the standard accuracy degrades compared to the training on the entire training set. One can acquire more accuracy at the cost of robustness by incorporating more problematic data into the training. This in fact introduces a new trade-off hyper-parameter $k$, namely the number of high-quality examples in the training set. % Compared with commonly used trade-off hyper-parameter $\lambda$, which trade the adversarial loss for standard loss in the loss function, and the perturbation radius $\varepsilon$, 
% Compared to other trade-off hyper-parameters such as the perturbation radius $\varepsilon$, $k$ extends the current performance of adversarial training methods to the far robustness side, as shown in Figure~\ref{fig:method-tradeoff}. For future work, this new trade-off curve can serve as a performance baseline.
% % It's also possible to extend the robustness side by more iterations (presumably based on FGSM's improved standard accuracy), but of course is computationally heavy

% \begin{figure*}[!ht]
% \centering
% \includegraphics[width=0.75\textwidth]{figure/method-tradeoff.pdf}
% \caption{}
% \label{fig:method-tradeoff}
% \end{figure*}

\begin{table}[h]
  \small
  \caption{Performance of adversarially training a pre-activation ResNet-18 on CIFAR-10 using PGD and TRADES with different training perturbation radii. % ``All'' denotes the adversarial training on the entire training set of CIFAR-10, while ``Friendly'' denotes the adversarial training on a high-quality training subset consisting of the least problematic examples.
  }
  \vspace{0.5ex}
  \label{table:remove-hurt}
  \centering
  \small
  \begin{tabular}{rcccc}
    \toprule
    Method & $\varepsilon~(/255)$ & \specialcell{\# Training \\ samples \\ removed (\%)} & Standard Acc (\%) & Robust Acc (\%)\\
    \midrule
    PGD & 8 & $0$ & $83.92$ & $48.12$\\
    PGD & 8 & $20\%$ & $82.26$ & $48.24$\\
    \midrule
    PGD & 12 & $0$ & $76.08$ & $48.95$\\
    PGD & 12 & $20\%$ & $77.85$ & $49.22$\\
    \midrule
    TRADES & 8 & $0$ & $81.73$ & $48.18$\\
    TRADES & 8 & $20\%$ & $78.16$ & $49.35$\\
    \midrule
    TRADES & 12 & $0$ & $76.56$ & $45.40$\\
    TRADES & 12 & $20\%$ & $73.53$ & $46.85$\\
    \bottomrule
    % \tablefootnote{$^*$ indicates the best hyper-parameter searched.}
  \end{tabular}
\end{table}

\section{Experiment details}
\label{sect:method-detail}
We adopt the following setting for all experiments unless otherwise noted.

\smallsection{Robustness evaluation}
We consider the robustness against $\ell_\infty$ norm-bounded adversarial attack with perturbation radius $8/255$. Throughout the paper, the following attacks are employed.
\begin{itemize}[leftmargin=*]
    \item \textbf{AutoAttack}: This is currently the strongest adversarial attack to the best of our knowledge. No parameter setting is required.epo
    % Throughout the paper, two versions of AutoAttack are used. 
    % By default, we will use a custom version due to computational constraint.
    % \begin{itemize}[leftmargin=*,nosep]
    %     \item Standard version, which consists of $\text{APGD}_{\text{CE}}$~\citep{Croce2020ReliableEO}, $\text{APGD}_{\text{DLR}}$~\citep{Croce2020ReliableEO}, FAB~\citep{Croce2020MinimallyDA} and Square Attack~\citep{Andriushchenko2020SquareAA}. This is the default version introduced in~\citet{Croce2020ReliableEO}. We use this version to report the numeric results in tables.
    %     \item Custom version, which consists of $\text{APGD}_{\text{CE}}$ and $\text{APGD}_{\text{DLR}}$. This version is faster and consumes less computation resources, and is often sufficient since the FAB-T and Square Attack in the standard version rarely attack any new examples successfully based on our experiments\footnote{We occasionally find $1$ out of $10^4$ examples in the test set of CIFAR-10 will be attacked successfully by FAB-T or Square Attack while not being attacked successfully by $\text{APGD}_{\text{CE}}$ and $\text{APGD}_{\text{DLR}}$.}. We use this version to report the results in figures, which typically cover a number of experiments.
    % \end{itemize}
    \item \textbf{PGD-10}: PGD with $10$ attack iterations and step size fixed to $2/255$.
    \item \textbf{PGD-1000}: PGD with $1000$ attack iterations. The step size is fixed to $2/255$, which is the best value suggested by \citet{Croce2020ReliableEO}.
    \item \textbf{Square Attack:} This is a strong score-based black-box attack and is also built into the standard AutoAttack. We list it separately here to eliminate the possibility that the adaptive design in AutoAttack can be sensitive to the data. We adopt $1$ restart with maximally $5000$ queries.
    \item \textbf{RayS:} This is a strong decision-based black-box attack that only requires the final prediction of the model. We set the maximum number of queries to be $10000$.
    \item \textbf{Transfer attack:} We use PGD to adversarially train a WRN-28-10 on the CIFAR-10 training set. The model at the best checkpoint in terms of the robustness will then serve as a surrogate model to generate the adversarial attack.
\end{itemize}

% \chengyu{If remove the table 1, remove the first version correspondingly. Table 2 (gradient masking) is using the second version.}

% maybe should also use another evaluation such as MT (multi-target) (An Alternative Surrogate Loss for PGD-based Adversarial Testing), as aligned with \citet{Gowal2020UncoveringTL}

\smallsection{Adversary setting}
\label{sect:method-adversary}
We conduct adversarial training with $\ell_\infty$ norm-bounded perturbations. We employ standard PGD training and TRADES as base methods, and two sophisticated variants GAIRAT and MART.
% , which can yield state-of-the-art robustness~\citep{Gowal2020UncoveringTL}.  
We fix the perturbation radius to $8/255$ unless otherwise noted. The number of attack iterations is fixed as $10$, and the perturbation step size is fixed as $2/255$. We adopt early stopping~\citep{Rice2020OverfittingIA} as a default strategy and report the best robustness obtained throughout the training. Additional settings specific to each method are listed as follows.
\begin{itemize}[leftmargin=*]
    \item \textbf{TRADES:} The regularization hyperparameter is fixed as $1/\lambda = 6.0$ as recommended in the official implementation.
    \item \textbf{GAIRAT:} A sample-wise weighting based on the distance to the decision boundary is employed in GAIRAT. The distance to the decision boundary is approximated by the minimum number of iterations $\kappa(x, y)$ to generate adversarial perturbation that successfully fools the model in the inner maximization. The sample-wise weighting is given by
    $$
        w(x, y) = \frac{1 + \tanh(\lambda + 5\cdot(1 - 2\cdot\kappa(x, y)/K))}{2},
    $$
    where $K$ is the maximum number of iterations. We fix $\lambda=0$ throughout the training. $\omega(x, y)$ will be normalized to ensure $\sum_i^n \omega(x, y)/n = 1$, where $n$ is the total number of training examples.
    \item \textbf{MART:} We adopt the version that employ the boosted cross entropy (BCE) loss in the outer minimization. The regularization hyperparameter is fixed as $1/\lambda = 6.0$ to be aligned with TRADES.
\end{itemize}
 
\smallsection{Training setting}
 We employ SGD as the optimizer. The momentum and weight decay are set as $0.9$ and $0.0005$ respectively, which are aligned with the common practice. For Wide ResNet, we conduct the training for 120 epochs, with the learning rate starting at 0.1 and reduced by a factor 10 at epoch 100 and 110. For other neural architectures, we conduct the training for 160 epochs, with the learning rate starting at 0.1 and reduced by a factor of 10 at epoch 80 and 120. 

%  Regarding the learning rate scheduler, when presenting the problems in adversarial training including robustness-accuracy trade-off, robust overfitting and robustness overestimation\footnote{\note{Figure \ref{}}}, we refer to \citet{Rice2020OverfittingIA} and conduct the training for $160$ epochs, with the learning rate starting at $0.1$ and reduced by a factor of $10$ at epoch $80$ and $120$. % When presenting the robustness improvement brought by our data-removal strategy\footnote{\note{Figure \ref{}}}, we align the setting with \citet{pang2021bag} and \note{conduct the training for $110$ epochs, with the learning rate starting at $0.1$ and reduced by a factor $10$ at epoch $100$ and $105$}.

\smallsection{Dataset}
We conduct experiments on three datasets including CIFAR-10, CIFAR-100 and Tiny-ImageNet without additional data.

\smallsection{Neural architecture}
We conduct experiments with pre-activation ResNet-18, VGG and Wide ResNet.

\smallsection{Hardware}
We conduct experiments on NVIDIA Quadro RTX A6000..

\section{Explanation}

In this section, we move one step further to probe into the effect of data quality in adversarial training. In previous works, it has been proven that standard classifier and robust classifier learn fundamentally different features~\citep{Tsipras2019RobustnessMB}. Useful features prevail in the dataset, but are not necessarily robust and comprehensible to human~\citep{Ilyas2019AdversarialEA}. In light of such analyses of robust/non-robust features, we motivate from a feature learning perspective and try to uncover the potential mechanisms of how low-quality data is interconnected with the existing difficulties in adversarial training including robustness-accuracy trade-off, robust overfitting and robustness overestimation. Under the assumption that similar features will be recognized in a specific class of examples, we selectively conduct the adversarial training on either one or a few classes of examples. In this way, we can isolate the impact of each set of similar features and analyze the effects of low-quality data on the learning of features.

% \note{One popular explanation is that the standard classifier and robust classifier learn fundamentally different features~\citep{Tsipras2019RobustnessMB}. Predictive features prevail in the dataset, but are not necessarily robust and comprehensible to human~\citep{Ilyas2019AdversarialEA}. The adversarial training prevents the learning of non-robust features and thus damages the accuracy of the classifier.}

% Bascially two type of methods
% 1. Attack targeted to one specifc class in inner maxmization
% 2. Add low-quality data with only one class to high-quality data

\subsection{Robustness-accuracy trade-off}
\label{sect:explain-tradeoff}

% \note{Such differentiation depends on the degree of ambiguity of the examples. On intrinsically ambiguous examples, adversarial training might not be suitable because the adversary will easily generate useful features of other classes.} 

Towards understanding the correlation between robustness-accuracy trade-off and data quality, we conjecture that low-quality data will cause the loss of useful features in adversarial training.

As shown in Figure~\ref{fig:low-quality-examples}, low-quality examples are intrinsically ambiguous from a human perspective. Therefore, if the perturbation radius is relatively large, the adversarial attack may generate reasonable images of classes other than the true class. Indeed, as shown in \ref{fig:low-quality-examples}, the adversarial counterparts of low-quality examples catch salient characteristics of the classes that the model predicts. In contrast, the adversarial counterparts of high-quality examples are still explicit images of the true classes. Here, we adversarially attack an example using PGD-20 with perturbation radius $\varepsilon=16/255$ based on the best model obtained through a PGD training. The perturbation radius $16/255$ is exactly twice the perturbation radius commonly used in adversarial training, which implies that adversarial counterparts generated on the low-quality examples during the training might be marginal cases from a human perspective.

Based on the above observation, we suspect that the adversarial training might not be suitable for low-quality examples. It is explained in \citet{Ilyas2019AdversarialEA} that adversarial training works because the adversary can exploit non-robust features of classes other than the true class, thus forcing the model to rely only on the robust features of the true class. However, due to the ambiguity, such ``distracting'' features generated on a low-quality example might be 
% useful features of other classes. Therefore, when the model is forced to classify such ``distracting''features into the true class, the recognizability of other classes might be damaged. 
too prominent such that it overwhelms the regular features of other classes when it is forced to be classified into the true class of this example, thus significantly damage the recognizability of other classes.

% -------- \chengyu{How to quantify prominent? Vary epsilon?}

\begin{figure}[!ht]
  \centering
  \includegraphics[width=0.8\linewidth]{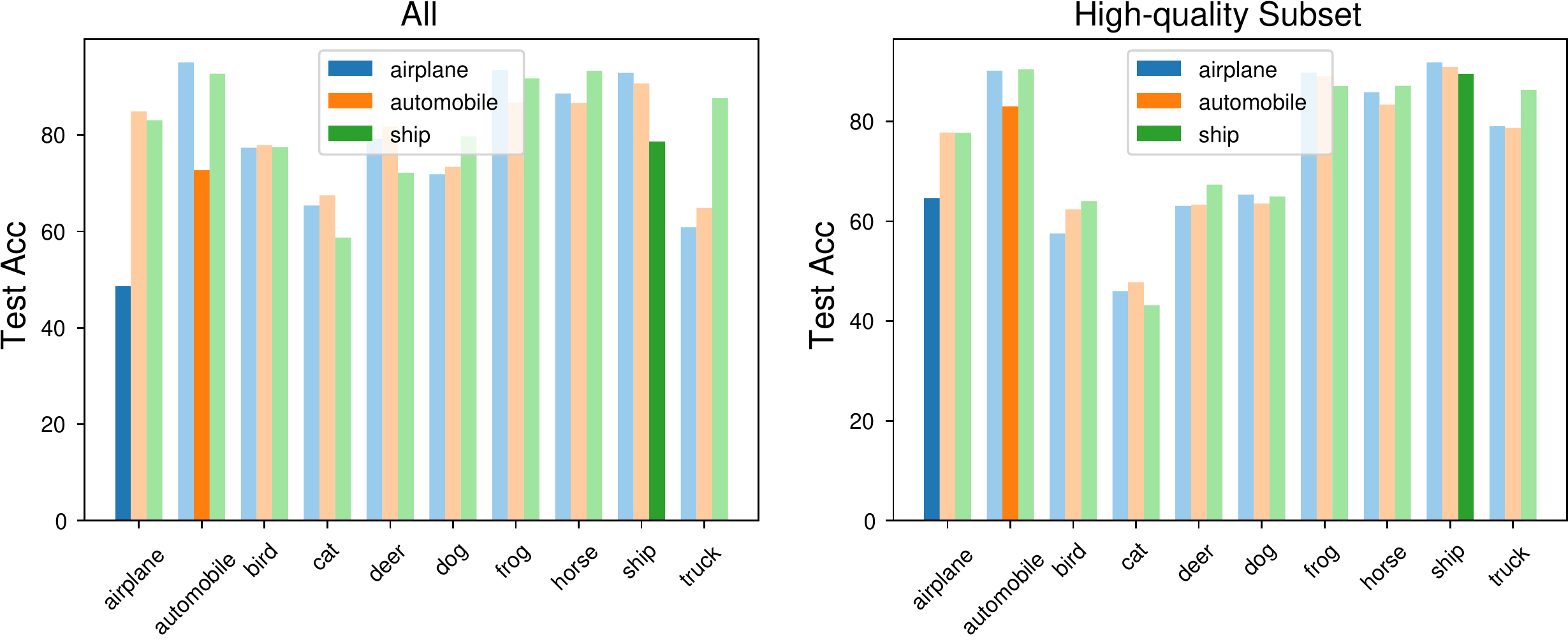}
  \caption{Fine tune the model obtained by standard training with adversarial examples generated by targeted attack on either all the training examples or the high-quality subset among them. When the attack in adversarial training is primarily targeted to ``Airplane'', the test accuracy of ``Airplane'' after fine-tuning is significantly lower, while the test accuracies of other classes are comparable. Instead, when the attack is targeted to ``Automobile'', the test accuracy of ``Automobile'' after fine-tuning is significantly lower. In contrast, when fine tuning the model only on the high-quality subset, such difference is not significant.}
\label{fig:tradeoff-targetattack}
\end{figure}

We manifest such loss of recognizability by robustly fine tuning the model obtained by standard training. Instead of untargeted attack, we use targeted attack in adversarial training\footnote{It is not common to use targeted attack in adversarial training, only for demonstration here.} to isolate the effect of recognizability loss. Specifically, for the PGD attack employed in adversarial training, we replace the inner maximization by a minimization towards a target class $c$, except for those examples have class $c$ as their true labels, where the target is directed to another class $c''$. We refer this special case as $c-c''$ adversarial training. Figure~ \ref{fig:tradeoff-targetattack} shows the average standard test accuracy produced by fine-tuning for 30 epochs using Airplane-Truck, Automobile-Truck and Ship-Truck adversarial training. These classes are selected because the model produces highest standard test accuracy on them. One can find that, when the attack in adversarial training is primarily targeted to ``Airplane'', the test accuracy of ``Airplane'' after fine-tuning is significantly lower, while the test accuracies of other classes are comparable. Instead, when the attack is targeted to ``Automobile'', the test accuracy of ``Automobile'' after fine-tuning is significantly lower. In contrast, when fine tuning the model only on the high-quality examples, such difference is not significant. This reflects the detail of how low-quality examples in adversarial training hurts the learning of useful features. % standard test accuracy. % In brief, when the attack is targeted to one certain class, it will impair the recognizability of this class.

% -------- \note{Add some discussions from geometric perspective, like the distance between classes and the adversary strength, since this is popular in mainstream works (``more data``, Fawzi?) (distance between low-quality data is small, therefore larger trade-off gap based on ``more data``)}

% \smallsection{Adversarial training on low-quality examples causes loss of robust recognizability}
\subsection{Robust overfitting}
\label{sect:explain-robustoverfitting}
% In previous section, we mentioned the low-quality examples may damage the recognizability in adversarial training. 
Here, we further show that the low-quality examples will also damage the robust recognizability. In adversarial training, the gradient-based adversary may generate robust features of another class on low-quality examples due to their intrinsic ambiguity, especially when the model is already relatively robust. Consequently, the model may forget 
% lose\note{forget}
the robust recognizability of other classes because it is forced to classify such robust feature to the original class, which leads to robust overfitting.

\begin{figure}[!ht]
  \centering
  \includegraphics[width=0.8\linewidth]{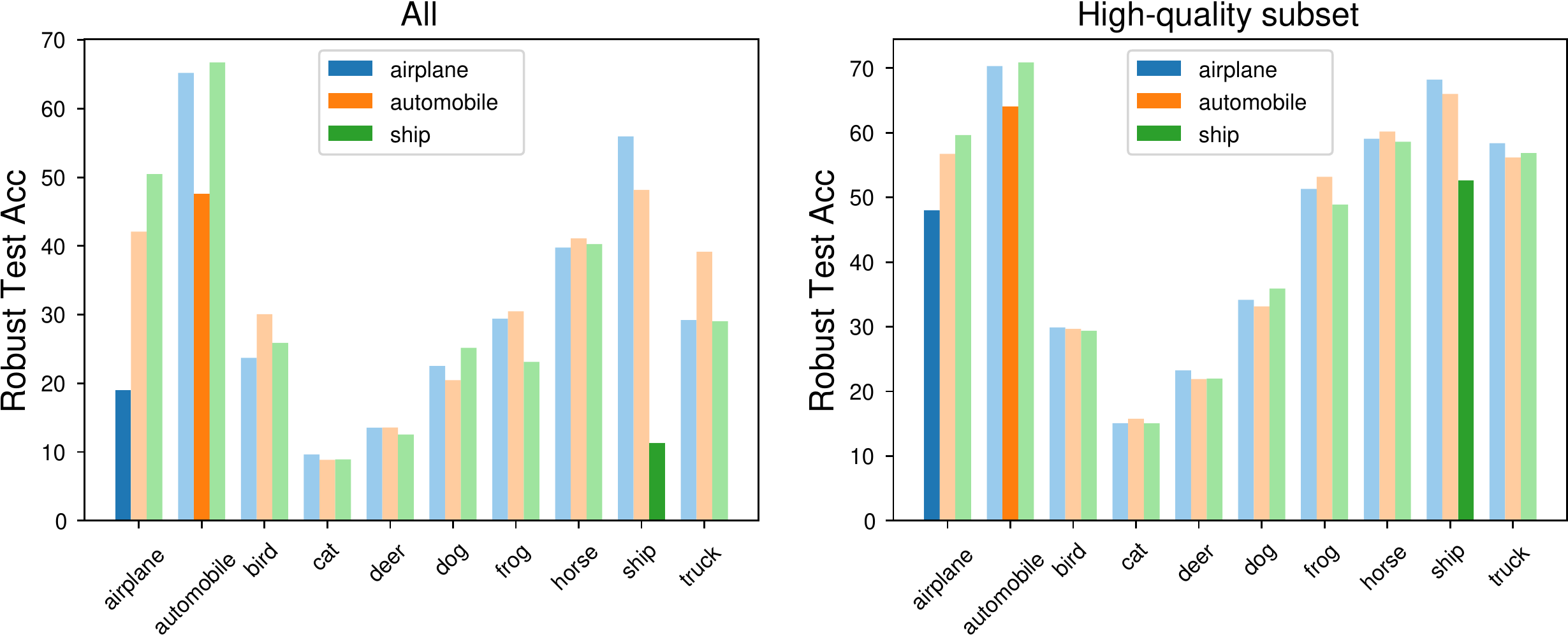}
  \caption{Fine tune the best model obtained in adversarial training with targeted attack on either all the training examples or the high-quality subset among them. When the attack in adversarial training is primarily targeted to ``Airplane'', the robust test accuracy of ``Airplane'' after fine-tuning is significantly lower, while the performance of other classes are comparable. We can observe similar patterns when the attack is targeted to ``Automobile'' or ``ship''. In contrast, when fine tuning the model only on the high-quality subset, the difference is  insignificant.}
\label{fig:explain-target-attack}
\end{figure}

We manifest the loss of recognizability by fine-tuning the best model obtained by a regular adversarial training. Similarly, we use targeted attack to isolate the effect. Figure~\ref{fig:explain-target-attack} shows the average robust test accuracy produced by fine-tuning the best model for 30 epochs using Airplane-Truck, Automobile-Truck and Ship-Truck adversarial training. These classes are selected because the model produces highest robust test accuracy on them. One can find similar results that, when the attack in adversarial training is primarily targeted to ``Airplane'', the robust test accuracy of ``Airplane'' after fine-tuning is significantly lower. Instead, when the attack is targeted to ``Automobile'', the robust test accuracy of ``Automobile'' after fine-tuning is significantly lower. In contrast, when fine tuning the model only on the high-quality data, such difference is not prominent.  % In brief, when the attack is targeted to one certain class, it will impair the recognizability of this class.

% \chengyu{This might not be correct based on the newest result (optimal eps=12), considering delete it} - future work
% In addition, since the capability of the adversary to generate meaningful features of another class is constrained by the perturbation radius $\epsilon$, we vary $\epsilon$ from $2/255$ to $32/255$ in the adversarial training, and check the best test accuracy can be obtained. Fig. \ref{fig:attackability-eps} shows that, in terms of robust test accuracy, the optimal $\epsilon$ on a high-quality subset is around $12/255$, while the optimal $\epsilon$ on a low-quality subset is around $8/255$. This implies that the low-quality examples are more inclined to generate troublesome adversarial perturbations and ruin the recognizability of the model.

% \smallsection{Problematic data causes conflict in the inner maximization}
\subsection{Robustnesss overestimation}
\label{sect:explain-gradientmasking}

\begin{figure}[!ht]
  \centering
  \includegraphics[width=0.8\textwidth]{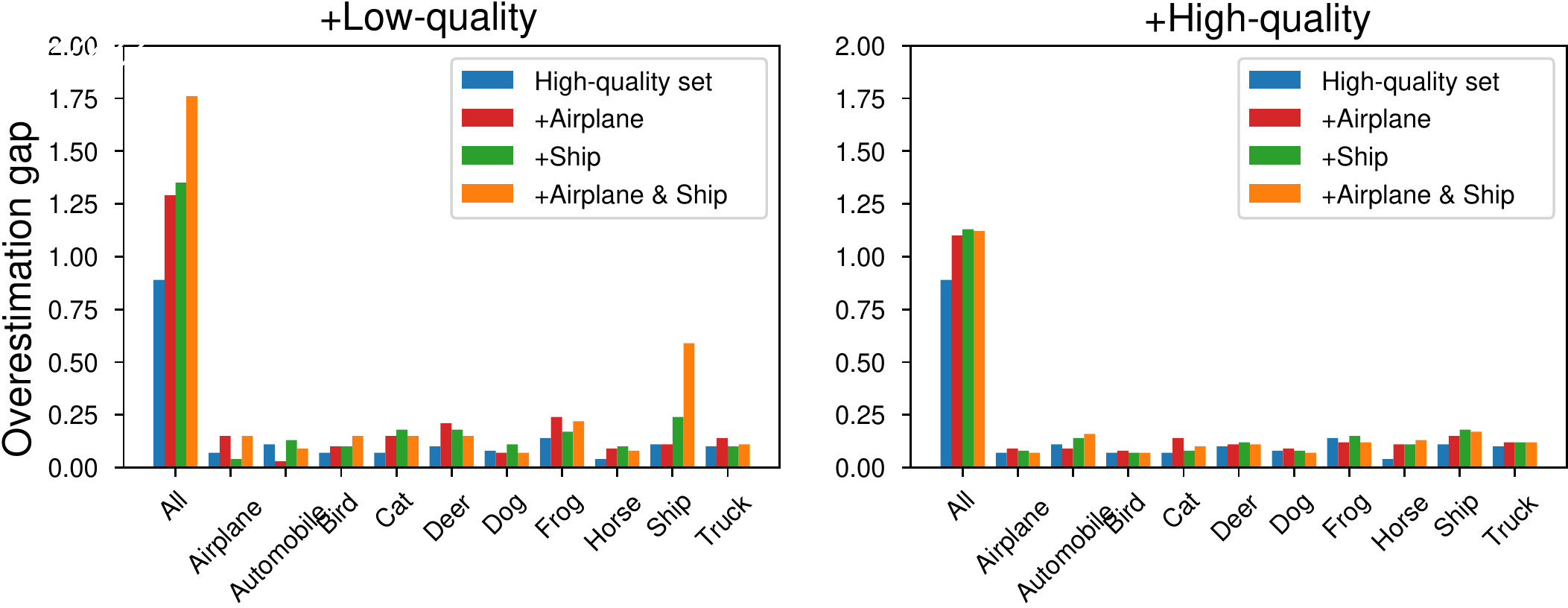}
  \caption{The overestimation gap, namely the difference between PGD-10 and Auto Attack evaluation, generated by adding examples of two competing classes into a high-quality subset.}
\label{fig:gradientmask-competing}
\end{figure}

\begin{figure}[!ht]
  \centering
  \includegraphics[width=0.8\textwidth]{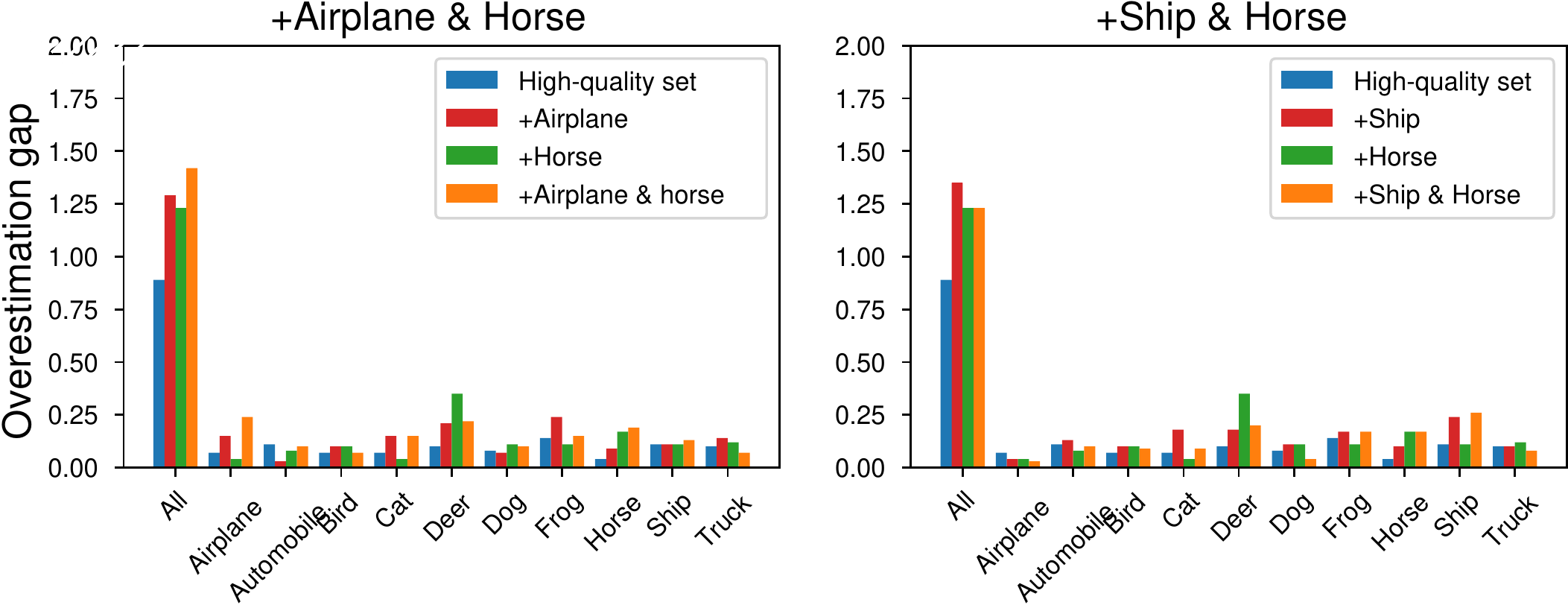}
  \caption{The overestimation gap generated by adding examples of two non-competing classes into a high-quality subset.}
\label{fig:gradientmask-noncompeting}
\end{figure}

We show that the low-quality data causes robustness overestimation through a mechanism which we refer as ``competing''. In Figure~~\ref{fig:gradientmask-competing}, we sample a subset by adding 500 additional low-quality examples\footnote{10\% of all the training examples in one class} to a class-balanced high-quality subset of size $10^4$, adversarially train the model on it, and show the overestimation gap. One can find that when we only add examples of one class either ``Airplane'' or ``Ship'', the overestimation gap increases, but not significantly compared to the overestimation gap of the original high-quality subset. However, if we add the examples of two ``competing'' classes ``Airplane'' and ``Ship'' at the same time, the overestimation gap increases substantially, and mostly attributes to the class ``Ship''. Here, ``competing'' classes means these two classes contain images that are likely to have similar features\footnote{One can refer to the sample images we showed in Figure~\ref{fig:low-quality-examples}}. The overestimation gap is not significant if the additional examples are not from two ``competing'' classes, as shown in Figure~\ref{fig:gradientmask-noncompeting} where we add additional examples from either ``Airplane'' and ``Horse'', or ``Ship'' and ``Horse''. The overestimation gap is also not significant if the additional examples are not low-quality, as shown in Figure~\ref{fig:gradientmask-competing} where we add additional high-quality examples, even if they are from competing classes.

 % major difference to previous: Inner maximization, Previous: Out minimization (assuming distracting feature is perfectly generated in the inner maximization)
Towards understanding this mechanism, we focus on the inner maximization since overestimation mainly indicates weakened generation of adversarial perturbation, while previously we assume this process is ideal. Recall that adversarial attack works because in the inner maximization, the adversary optimizes towards classes other than the original class and thus can exploit distracting features to fool the model~\citep{Ilyas2019AdversarialEA}. However, since the ambiguous low-quality examples from different classes tend to contain similar features, the adversary may optimize towards different classes starting from similar features, subsequently damage its capability to exploit distracting features of these classes.

\end{document}